\documentclass{article} 
\usepackage{iclr2025_conference,times}

\usepackage{wrapfig}
\usepackage{lipsum}
\usepackage{wrapfig}

\usepackage{booktabs}
\usepackage{multirow}
\usepackage{amssymb}
\usepackage{graphicx}
\usepackage{colortbl}
\usepackage{xcolor}

\usepackage{hyperref}

\usepackage{url}
\usepackage{multicol}
\usepackage{bm}
\usepackage{aliascnt}
\usepackage{adjustbox}
\usepackage{siunitx}
\usepackage{booktabs}
\usepackage{multirow}
\usepackage{enumitem}
\usepackage{booktabs}
\usepackage{amssymb}
\usepackage{amsmath}
\usepackage{graphicx}
\usepackage{longtable}
\usepackage{graphicx}
\usepackage{subcaption}
\usepackage{calc}
\usepackage[most]{tcolorbox}
\usepackage{listings}
\usepackage{wrapfig}
\usepackage{caption}

\definecolor{uclablue}{rgb}{0.15, 0.45, 0.68}
\hypersetup{
    breaklinks,
    citecolor=uclablue,
    colorlinks=true,
}

\newtcolorbox{AIbox}[2][]{aibox,title=#2,#1}
\tcbset{
  aibox/.style={
    top=10pt,
    colframe=black,
    colbacktitle=black,
    coltitle=white,
    center,
  }
}

\lstdefinelanguage{prompt}{
    basicstyle=\scriptsize\ttfamily,
    mathescape=true,
    escapebegin=\color{latentcolor},
    escapeend={},
    escapechar=@,
    stringstyle = \color{myorange},
    showstringspaces = false,
    moredelim = [s][\color{mypink}]{`}{`},
    moredelim = [s][\color{mybrown}]{```json}{```},
    moredelim = [s][\color{latentcolor}]{<StartOfLatent>}{<EndOfLatent>},
    literate = %
        {\ \ a.\ }{{\textcolor{mypurple}{\ \ a.\ }}}5
        {\ \ b.\ }{{\textcolor{mypurple}{\ \ b.\ }}}5
        {\ \ c.\ }{{\textcolor{mypurple}{\ \ c.\ }}}5
        {\ \ d.\ }{{\textcolor{mypurple}{\ \ d.\ }}}5
        {\ \ e.\ }{{\textcolor{mypurple}{\ \ e.\ }}}5
        {\ \ f.\ }{{\textcolor{mypurple}{\ \ f.\ }}}5
        {\ \ g.\ }{{\textcolor{mypurple}{\ \ g.\ }}}5
        {\ \ h.\ }{{\textcolor{mypurple}{\ \ h.\ }}}5
        {\ I.\ }{{\textcolor{mypurple}{\ I.\ }}}4
        {\ II.\ }{{\textcolor{mypurple}{\ II.\ }}}5
        {\ III.\ }{{\textcolor{mypurple}{\ III.\ }}}6
        {\ IV.\ }{{\textcolor{mypurple}{\ IV.\ }}}5
        {\ V.\ }{{\textcolor{mypurple}{\ V.\ }}}4
}

\newtcblisting[auto counter, number within=subsection]{prompt}[4][]{%
  before skip=6pt, after skip=6pt,
  top=6pt, bottom=6pt, left=7pt, right=7pt,
  enhanced, breakable,
  colback=white, colframe=gray!65,
  boxrule=0.6pt, arc=2pt,
  drop shadow={black!15},
  listing only,
  listing options={
    language=prompt,
    upquote=true,
    basicstyle=\scriptsize\ttfamily \setlength{\baselineskip}{1.1\baselineskip},
    columns=fullflexible,
    breaklines=true,
    breakindent=0pt,
    xleftmargin=\ifx\empty#2\empty0pt\else#2\fi,
    xrightmargin=\ifx\empty#3\empty0pt\else#3\fi,
    aboveskip=0pt,
    belowskip=0pt,
    showstringspaces=false,
  },
  title={\ifstrempty{#4}{Prompt \thetcbcounter}{Prompt \thetcbcounter: #4}},
  colbacktitle=gray!6, coltitle=black,
  attach boxed title to top left={yshift=-1mm, xshift=6pt},
  boxed title style={colback=gray!6, boxrule=0pt, sharp corners},
  #1
}

\newtcblisting[auto counter, number within=subsection]{showcase}[4][]{%
  before=\par\vspace{\baselineskip},
  after=\par,
  width=\linewidth,
  enhanced,
  arc=0em,
  boxrule=1pt,
  listing only,
  listing options={
    language=prompt,
    upquote=true,
    basicstyle=\scriptsize\ttfamily \setlength{\baselineskip}{1.1\baselineskip},
    breaklines=true,
    breakindent=0pt,
    xleftmargin=\ifx\empty#2\empty-12pt\else#2\fi,
    xrightmargin=\ifx\empty#3\empty-5pt\else#3\fi,
    aboveskip=-4pt,
    belowskip=-4pt,
    columns=fullflexible,
    },
  colback=white,
  colframe=gray,
  colbacktitle=gray!5,
  coltitle=black,
  attach boxed title to top center={yshift=-3mm},
  box align=center,
  parbox=false,
  title={\ifstrempty{#4}{Example \thetcbcounter}{Example \thetcbcounter: #4}},
  #1
}

\definecolor{linkColor}{rgb}{0.2,0.4,0.6}
\definecolor{myblue}{HTML}{0379AC}
\definecolor{myred}{HTML}{A50E50}
\definecolor{myorange}{RGB}{238, 133, 74}
\definecolor{mypink}{HTML}{D7388C}
\definecolor{mybrown}{HTML}{8B4513}
\definecolor{mypurple}{HTML}{6A0DAD}
\definecolor{latentcolor}{named}{cyan}
\definecolor{normalcolor}{RGB}{0, 0, 0}
\usepackage{marvosym}
\definecolor{lightblue1}{rgb}{0.97, 0.985, 1}
\definecolor{lightblue2}{rgb}{0.92, 0.965, 1}
\definecolor{lightblue3}{rgb}{0.84, 0.93, 1}
\definecolor{lightblue4}{rgb}{0.74, 0.87, 1}
\definecolor{lightblue5}{rgb}{0.64, 0.81, 1}
\definecolor{lightblue6}{rgb}{0.54, 0.75, 1}

\definecolor{lightgreen1}{rgb}{0.97, 1.00, 0.97}
\definecolor{lightgreen2}{rgb}{0.92, 0.98, 0.92}
\definecolor{lightgreen3}{rgb}{0.84, 0.95, 0.84}
\definecolor{lightgreen4}{rgb}{0.74, 0.91, 0.74}
\definecolor{lightgreen5}{rgb}{0.64, 0.86, 0.64}
\definecolor{lightgreen6}{rgb}{0.54, 0.81, 0.54}

\definecolor{lightorange1}{rgb}{1.00, 0.98, 0.95}
\definecolor{lightorange2}{rgb}{1.00, 0.95, 0.85}
\definecolor{lightorange3}{rgb}{1.00, 0.90, 0.70}
\definecolor{lightorange4}{rgb}{1.00, 0.85, 0.55}
\definecolor{lightorange5}{rgb}{1.00, 0.80, 0.40}
\definecolor{lightorange6}{rgb}{1.00, 0.75, 0.30}

\definecolor{lightpurple1}{rgb}{0.985, 0.97, 1.00}
\definecolor{lightpurple2}{rgb}{0.96, 0.92, 1.00}
\definecolor{lightpurple3}{rgb}{0.93, 0.84, 1.00}
\definecolor{lightpurple4}{rgb}{0.87, 0.74, 1.00}
\definecolor{lightpurple5}{rgb}{0.81, 0.64, 1.00}
\definecolor{lightpurple6}{rgb}{0.75, 0.54, 1.00}

\definecolor{lightred1}{rgb}{1.00, 0.97, 0.97}
\definecolor{lightred2}{rgb}{1.00, 0.92, 0.92}
\definecolor{lightred3}{rgb}{1.00, 0.84, 0.84}
\definecolor{lightred4}{rgb}{1.00, 0.74, 0.74}
\definecolor{lightred5}{rgb}{1.00, 0.64, 0.64}
\definecolor{lightred6}{rgb}{1.00, 0.54, 0.54}

\definecolor{lightcyan1}{rgb}{0.97, 1.00, 1.00}
\definecolor{lightcyan2}{rgb}{0.92, 0.98, 0.98}
\definecolor{lightcyan3}{rgb}{0.84, 0.95, 0.96}
\definecolor{lightcyan4}{rgb}{0.74, 0.91, 0.94}
\definecolor{lightcyan5}{rgb}{0.64, 0.87, 0.92}
\definecolor{lightcyan6}{rgb}{0.54, 0.83, 0.90}

\usepackage{xcolor,colortbl}
\definecolor{Gray}{gray}{0.85}
\definecolor{LightCyan}{rgb}{0.88,1,1}
\definecolor{greyC}{RGB}{180,180,180}
\definecolor{greyL}{RGB}{235,235,235}
\definecolor{citeColor}{RGB}{0,20,115}
\hypersetup{colorlinks,linkcolor={red},citecolor={citeColor},urlcolor={citeColor}}
\definecolor{shadecolor}{rgb}{0.92,0.92,0.92}

\usepackage{graphicx}
\usepackage{subcaption}
\usepackage{url}
\usepackage{booktabs}
\usepackage{multirow}
\usepackage{cleveref}
\crefname{template}{Template}{Template}

\usepackage{enumitem}
\usepackage[most]{tcolorbox}
\usepackage{pifont}
\definecolor{rliableblue}{RGB}{0, 102, 204}

\definecolor{modelcolor}{HTML}{1F4E79}
\usepackage{xspace}
\newcommand{\modelname}{\textsc{\textcolor{modelcolor}{Hydra-X}}\xspace}
\newcommand{\tokname}{\textsc{Hydra-XTok}\xspace}

\providecommand{\g}[1]{\textcolor{black!50}{#1}}

\tcbset{
    myblueboxstyle/.style={
        colback=lightblue3,
        colframe=lightblue5,
        coltitle=black,
        fonttitle=\bfseries,
        boxrule=1pt,
        width=\linewidth,
        left=2pt,
        right=2pt,
        top=2pt,
        bottom=2pt,
        sharp corners,
        boxsep=2pt
    }
}

\newtcolorbox{finding}[1][]{%
  enhanced, breakable,
  colback=blue!3, colframe=blue!50!black,
  arc=3pt, boxrule=0.6pt,
  left=8pt, right=8pt, top=4pt, bottom=4pt,
  fonttitle=\bfseries,
  title={#1}}

\newtcolorbox{keyfindingsbox}{%
    enhanced, breakable,
    colback=white!95!gray, colframe=modelcolor,
    arc=3pt, boxrule=0.6pt,
    left=8pt, right=8pt, top=4pt, bottom=4pt
}

\lstdefinestyle{iclrstyle}{
    language=Python,
    basicstyle=\ttfamily\small,
    columns=fullflexible,
    keepspaces=true,
    showspaces=false,
    showstringspaces=false,
    commentstyle=\color{gray}\itshape,
    keywordstyle=\color{codekw}\bfseries,
    stringstyle=\color{myorange},
    escapechar=|,
    frame=none,
    xleftmargin=1.5em,
    aboveskip=0.5em,
    belowskip=0.5em,
    breaklines=true,
    breakindent=0pt,
}

\usepackage{algorithm}
\usepackage{algpseudocode}
\usepackage{listings}
\usepackage{xcolor}
\usepackage{etoolbox}
\makeatletter
\AfterEndEnvironment{algorithm}{\let\@algcomment\relax}
\AtEndEnvironment{algorithm}{\kern2pt\hrule\relax\vskip3pt\@algcomment}
\let\@algcomment\relax
\newcommand\algcomment[1]{\def\@algcomment{\footnotesize#1}}
\renewcommand\fs@ruled{\def\@fs@cfont{\bfseries}\let\@fs@capt\floatc@ruled
  \def\@fs@pre{\hrule height.8pt depth0pt \kern2pt}%
  \def\@fs@post{}%
  \def\@fs@mid{\kern2pt\hrule\kern2pt}%
  \let\@fs@iftopcapt\iftrue}
\makeatother

\NewDocumentCommand{\xx}
{ mO{} }{\textcolor{blue}{\textsuperscript{\textit{todo}}\textsf{\textbf{\small[#1]}}}}

\usepackage{soul}
\definecolor{codeblue}{rgb}{0.25,0.5,0.5}
\definecolor{codekw}{rgb}{0.85, 0.18, 0.50}
\definecolor{diffgreen}{rgb}{0.0, 0.6, 0.0}
\definecolor{diffred}{rgb}{0.8, 0.0, 0.0}

\title{{\modelname}: Native Unified Multimodal Models with \\ Holistic Visual Tokenizers}

\author{
Guozhen Zhang$^{1,\dagger,*}$,
Xuerui Qiu$^{2,4,\dagger,*}$,
Yutao Cui$^{3,\dagger}$, \\
Tianhui Song$^{3}$,
Changlin Li$^{3}$,
Junzhe Li$^{3}$,
Tao Huang$^{3}$,
Xiao Zhang$^{3}$,
Yang Li$^{3}$,
Jianbing Wu$^{3}$, \\
Miles Yang$^{3}$,
Zhao Zhong$^{3}$,
Liefeng Bo$^{3}$,
Limin Wang$^{1,5,\ddagger}$\\
\textbf{$^1$Nanjing University}  \textbf{$^2$CASIA}  \textbf{$^3$Tencent Hunyuan}\\
\textbf{$^4$Zhongguancun Academy}  \textbf{$^5$Shanghai AI Lab}\\
\texttt{zgzaacm@gmail.com} \quad \texttt{lmwang@nju.edu.cn}
}

\begin{document}
\maketitle
\renewcommand*{\thefootnote}{\fnsymbol{footnote}}
\footnotetext{$*$ Work done during internship at Tencent Hunyuan. $\dagger$ Equal contribution. $\ddagger$ Corresponding author.}

\begin{abstract}
Holistic visual tokenizers are fundamental to unified multimodal models (UMMs) as they map diverse visual inputs into a unified representation space.
In this paper, we present \modelname, the first UMM that unifies image and video tokenization within a single Vision Transformer (ViT). Our design is driven by two core challenges: efficiently injecting spatiotemporal reconstruction capability into a native ViT, and embedding image- and video-level semantic awareness into the latent space. To address the first, comprehensive ablations reveal two key findings: \textbf{(1)} frame-level causal temporal attention suffices for visual reconstruction, whereas full spatiotemporal attention degrades it; and \textbf{(2)} hierarchical temporal compression substantially outperforms single-step alternatives. To tackle the second, we propose a lightweight decompressor that upsamples temporally compressed features under joint image-video teacher supervision, thereby enforcing complementary semantic structures within the compact latent space. Building on this holistic tokenizer, we further propose a principled improvement of the editing pipeline: source-target interaction should occur at the latent level inside the tokenizer rather than at the semantic level inside the LLM, substantially improving editing consistency and accelerating convergence. Instantiated at the 7B dense model, \modelname achieves strong performance across image and video understanding and generation tasks, paving the way for future unified-tokenizer UMMs.
\end{abstract}

\section{Introduction}
\label{sec:intro}

Unified multimodal models (UMMs)~\citep{xie2025show, liu2025tuna, bagel, hydra, zhou2024transfusion} have recently emerged as a powerful paradigm that jointly trains a single autoregressive backbone for both visual understanding and generation. A central design choice is how visual inputs are encoded: existing systems either deploy \emph{decoupled visual encoders} that pair a ViT encoder with a separate VAE encoder for the two tasks~\citep{bagel,ma2025janusflow,zhou2024transfusion}, or adopt a \emph{unified visual tokenizer} that maps diverse visual inputs into a single representation space shared by both tasks~\citep{xie2025show,liu2025tuna,beyond2026,hydra,wu2025harmonizing,unitok}. The latter approach offers distinct architectural advantages: it eliminates the representational mismatch between heterogeneous encoders that the LLM must otherwise reconcile, and opens a pathway for the mutual reinforcement between understanding and generation.

\begin{figure}[!htbp]
  \centering
  \includegraphics[width=0.9\linewidth]{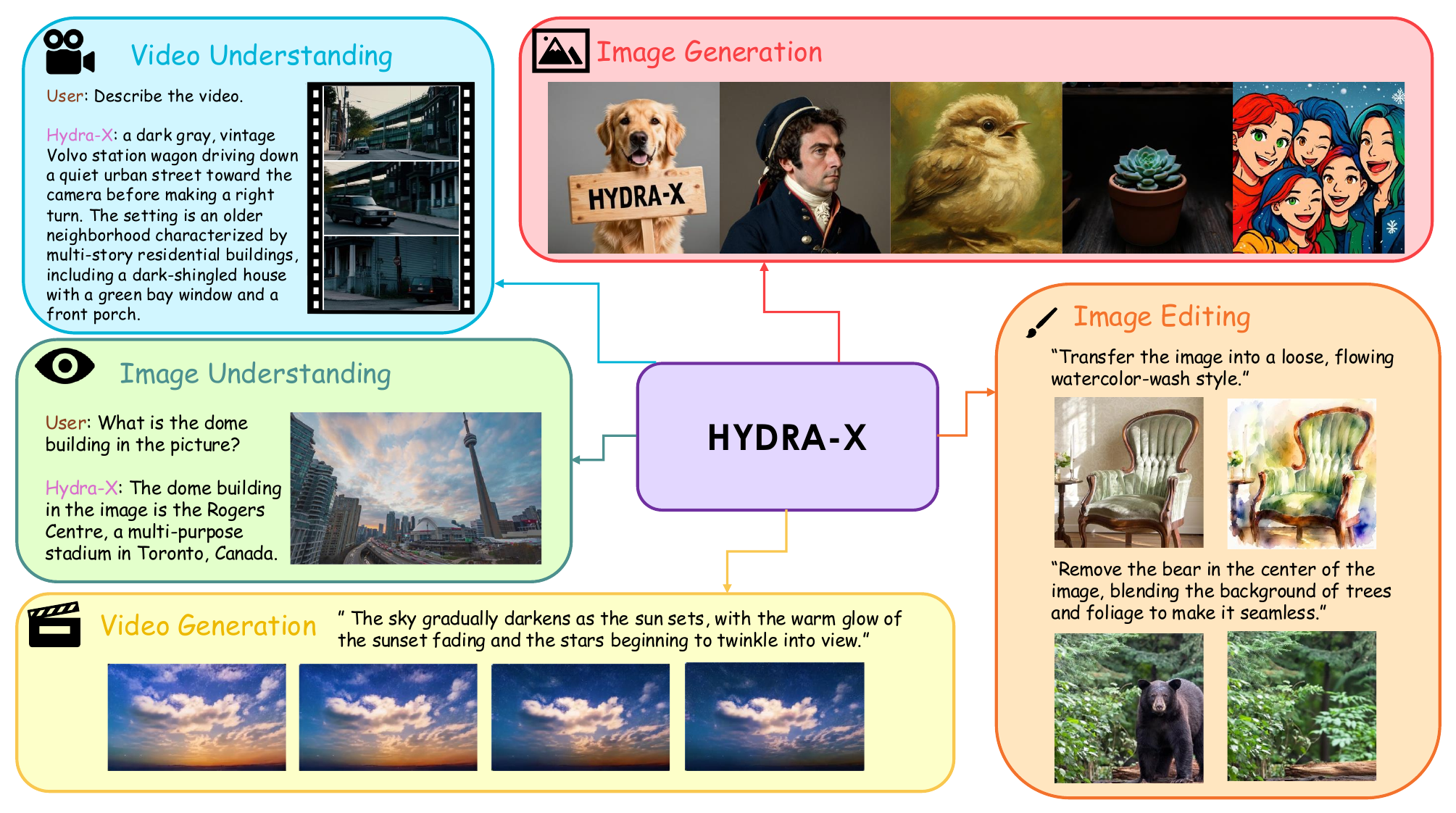}
  \caption{\modelname is a native UMM that unifies image/video understanding, image/video generation, and instruction-guided image editing through one holistic tokenizer \tokname.}
  \label{fig:teaser}
\end{figure}

\textbf{While unified tokenization has been extensively explored for static images, a \emph{holistic} tokenizer that binds images and videos into a single representation space has received much less attention.} Existing video-capable UMMs typically adopt one of two ad-hoc strategies. The first paradigm relies on frame-wise tokenizers that apply an image semantic encoder independently to each frame~\citep{beyond2026}. Without any temporal interaction inside the tokenizer, the resulting representation cannot capture cross-frame dynamics such as motion or short-horizon causality, leaving the downstream LLM with disjoint per-frame features that carry no inherent video structure. The second paradigm employs cascaded designs that stack a 3D causal VAE before a semantic encoder~\citep{xie2025show, liu2025tuna}. Although this packs the temporal axis, the VAE is trained in isolation without any semantic constraint, and may inadvertently discard information critical for understanding.

In this paper, we present \modelname, the first UMM framework built upon \tokname, a unified visual tokenizer that handles both image and video encoding within a single Vision Transformer (ViT). Our overall design follows the image-only UMM framework HYDRA~\citep{hydra}, which compresses intermediate-layer ViT features into a compact latent and then reconstructs semantic feature from it. Extending this paradigm to jointly support images and videos, however, raises two core challenges: \emph{(a)} efficiently injecting spatiotemporal reconstruction capability into a native ViT, and \emph{(b)} embedding both image- and video-level semantic awareness into the shared latent space.

Our investigation of the first challenge yields two findings that run counter to conventional intuition. \textbf{(1)} Although full spatiotemporal attention is the most natural extension to video, it actively degrades reconstruction by disrupting the locality and structure encoded during image pretraining. Surprisingly, frame-level causal temporal attention with a minimal temporal receptive field, attending only to the immediately preceding frame, comprehensively outperforms its global counterpart. \textbf{(2)} A single-step patchify substantially underperforms a hierarchical patchify that distributes temporal compression across multiple stages, indicating that the temporal axis benefits from progressive, multi-scale folding. Together, these two design choices enable \tokname to surpass the reconstruction fidelity of dedicated 3D-conv video VAEs such as Wan2.2-VAE~\citep{wan2.2}.

To address the second challenge, we extend the established paradigm of semantic distillation~\citep{hydra,wu2025harmonizing,unitok} from images to video, and uncover a fundamental asymmetry: while image latents can readily reuse existing semantic teachers, no available video encoder operates at the compressed temporal resolution of our latent, leaving the video stream without a natural source of semantic supervision. We resolve this asymmetry through a remarkably simple addition: a lightweight \emph{Decompressor} that lifts the compressed latent back to its native temporal length, enabling direct distillation from pretrained image and video teachers~\citep{tschannen2025siglip,internvideo} at full frame rate. Under this dual spatiotemporal supervision, the compact latent simultaneously preserves pixel-level fidelity and rich spatiotemporal semantic structure, substantially advancing both understanding and generation in UMMs.

Building on this holistic tokenizer, \modelname unifies five UMM tasks within a single shared encoder, as shown in Figure~\ref{fig:teaser}: image/video generation, image/video understanding, and image editing. Yet editing in particular exposes a fundamental flaw in both HYDRA and cascaded designs: by feeding the LLM only post-encoder semantic features, they confine source-target interaction to the semantic level and forfeit the fine-grained structural information that resides at the latent. To resolve this, we propose a principled inversion of the design: \tokname jointly tokenizes source and target with cross-frame interaction, fusing structural details directly into the target before reaching the LLM. This early latent-level interaction substantially improves editing consistency and accelerates convergence.

Instantiated at the 7B scale on top of Qwen2.5-7B-Instruct~\citep{qwen2.5}, \modelname achieves strong performance across image and video understanding and generation tasks. More importantly, it elevates the visual tokenizer from a specialized image-processing component to a holistic image-and-video interface, laying a solid foundation for future unified-tokenizer UMM exploration.

\section{Related Work}
\label{sec:related}

\subsection{Visual Tokenizers for Unified Multimodal Models}

A growing body of work unifies reconstruction and semantics within a single visual tokenizer. For images, RAE~\citep{rae,scale-rae-2026} freezes a semantic encoder and learns a pixel decoder, while several unified-tokenizer designs~\citep{yue2025uniflow,yao2025vtp,unitok,qu2025tokenflow,dualtoken,toklip,unilip} co-train reconstruction and understanding within a single ViT. HYDRA~\citep{hydra} introduces a progressive ViT with a Generation--Semantic Bottleneck for compress-then-restore semantic distillation, which \tokname inherits. Aligning generative latents with semantic features has further been shown to mutually benefit both tasks~\citep{wang2024reconstructive,repa,vavae,ma2025genhancer,wang2025autoregressive,Wang_2023_CVPR}. Joint image-and-video tokenization, however, remains largely under-explored: for video, 3D-convolutional VAEs~\citep{magvit2,wan2.2} dominate but lack any semantic structure. A recent work, AToken~\citep{atoken}, unifies images and videos within a single tokenizer for reconstruction and understanding, but emits task-specific output features for the two objectives and therefore does not yield a unified representation. To our knowledge, \modelname is the first UMM framework to unify image and video within a single ViT-based tokenizer, augmenting HYDRA's philosophy with explicit temporal causality, hierarchical patchify, and a Decompressor for spatiotemporal semantic awareness.

\subsection{Native Unified Multimodal Models}

UMMs aim to handle visual understanding and generation within a single backbone, and existing systems can be broadly grouped into three families that differ in how tightly the two objectives share parameters and representations. \textit{Composite} UMMs~\citep{metamorph,blip3-o,lin2025uniworld,pan2025transfer,unilip} bridge pretrained understanding and generation models via lightweight adapters or projection layers; this preserves the strengths of each specialised model but leaves the synergy between the two tasks shallow, as gradients rarely flow across the modality boundary and the two backbones never see a shared latent. \textit{Native} UMMs instead train both objectives jointly from the start, and further bifurcate by their choice of visual representation. Quantised-token approaches~\citep{team2024chameleon,showo,emu3,zhou2024transfusion} cast visual generation as next-token prediction over a VQ codebook, which unifies the LLM interface but inherits the reconstruction loss and codebook-collapse pathologies of VQ tokenizers, capping the achievable visual fidelity. Decoupled designs~\citep{ma2025janusflow,janus,chen2025janus,bagel,mogao,li2025onecat,fan2025unified} side-step this ceiling by routing understanding through a semantic encoder and generation through a separately trained VAE; the price is a duplicated visual pathway whose two streams compete for LLM attention and whose representations must be re-aligned downstream. The most recent line we build on are \textit{unified-encoder} UMMs such as TransNext~\citep{beyond2026}, Show-o2~\citep{xie2025show}, and TUNA~\citep{liu2025tuna}, which share a single visual tokenizer across both tasks and recover the architectural cleanliness of composite systems while retaining joint optimisation. We extend this line in two directions: from images to a unified image-and-video tokenizer, and from independent per-input encoding to a tokenizer-stage source--target interaction tailored for editing.

\subsection{Image Editing in Unified Multimodal Models}

Image editing is the canonical task in which a UMM must condition the target image on a structurally similar source image, and existing pipelines differ mainly in where this conditioning is injected. The first family relies on dedicated condition adapters: ControlNet-style branches~\citep{zhang2023controlnet} attach a parallel encoder that injects spatially aligned source features into the generator, while reference-token streams as used in BAGEL~\citep{bagel} prepend the source as an extra context that the LLM attends to. Both families add either parameters or context length, and the source representation is shaped specifically for the generation head rather than shared with the understanding side. Closer to our setting, the unified-encoder UMMs Show-o2~\citep{xie2025show} and TUNA~\citep{liu2025tuna} reuse a single tokenizer for both the source and the target, but still encode the two images \emph{independently}; only their post-encoder semantic features are concatenated at the LLM input, so any cross-image alignment must be reconstructed by the LLM from two already-compressed semantic streams, with the fine-grained pre-bottleneck structure inaccessible. We instead place the source and target in the same temporal window of \tokname and process them in a single forward pass, allowing source--target interaction to begin at the latent level inside the tokenizer's causal Sem-ViT and propagate before reaching the LLM. This reuses the temporal pathway already trained for video, removes any extra cross-image attention module, and exposes the LLM to a target representation that has already absorbed source structure.

\section{Preliminaries: Representation-Harmonized Tokenization}
\label{sec:background}

Our overall design follows the image-only UMM framework HYDRA~\citep{hydra}. At its core is a single ViT split into a \emph{Gen-ViT} and a \emph{Sem-ViT}, connected by a \emph{Generation--Semantic Bottleneck} that supports generation and semantic perception within one backbone. Given an input image $\mathbf{x}\!\in\!\mathbb{R}^{H\times W\times 3}$, the Gen-ViT first produces a feature $\mathbf{h}$ rich in structural primitives, which the Bottleneck projects into a compact latent $\mathbf{z}\!\in\!\mathbb{R}^{N\times C}$ suitable for generation. The Sem-ViT then un-projects $\mathbf{z}$ back into a high-dimensional semantic feature $\mathbf{s}$, which is aligned with a pretrained semantic teacher $\mathcal{T}$ via distillation:
\begin{equation}
  \mathbf{x} \;\xrightarrow{\;\text{Gen-ViT}\;}\;
  \mathbf{h} \;\xrightarrow{\;\text{Bottleneck}\;}\;
  \mathbf{z} \;\xrightarrow{\;\text{Sem-ViT}\;}\;
  \mathbf{s} \;\xleftarrow{\;\text{align}\;}\;\mathcal{T}(\mathbf{x}).
  \label{eq:hydra-pipeline}
\end{equation}
The downstream LLM operates exclusively on the Sem-ViT output $\mathbf{s}$ for both understanding and generation, whereas the pixel decoder that reconstructs images from $\mathbf{z}$ is invoked only during tokenizer training. We retain this overall design and extend it from images to videos through explicit temporal causality, hierarchical patchify, and a Decompressor introduced in Section~\ref{sec:tokenizer}.

\section{\tokname: Holistic Visual Tokenization in a Single ViT}
\label{sec:tokenizer}

\tokname is designed as the visual interface of \modelname: before any token reaches the LLM, it must be compact enough for generation, faithful enough for reconstruction, and semantic enough for understanding. We initialize Gen-ViT and Sem-ViT from SigLIP\,2~\citep{tschannen2025siglip}; all UMM-side ablations use Qwen2.5-1.5B~\citep{qwen2.5}. The tokenizer is trained with a reconstruction term and two semantic distillation terms:
\begin{equation}
  \mathcal{L}_{\tokname}
  = \mathcal{L}_{\text{rec}}
  + \lambda\mathcal{L}_{\text{dist}},
  \label{eq:tokenizer-obj}
\end{equation}
where $\mathcal{L}_{\text{rec}}$ keeps the compact latent pixel-faithful, $\mathcal{L}_{\text{dist}}$ aligns Sem-ViT features with semantic features. Detailed recipes are in Appendix~\ref{tokenizer_loss}.

\subsection{Spatiotemporal Reconstruction in a ViT}
\label{sec:tokenizer:rec}

\begin{figure}[t]
  \centering
  \includegraphics[width=0.9\linewidth]{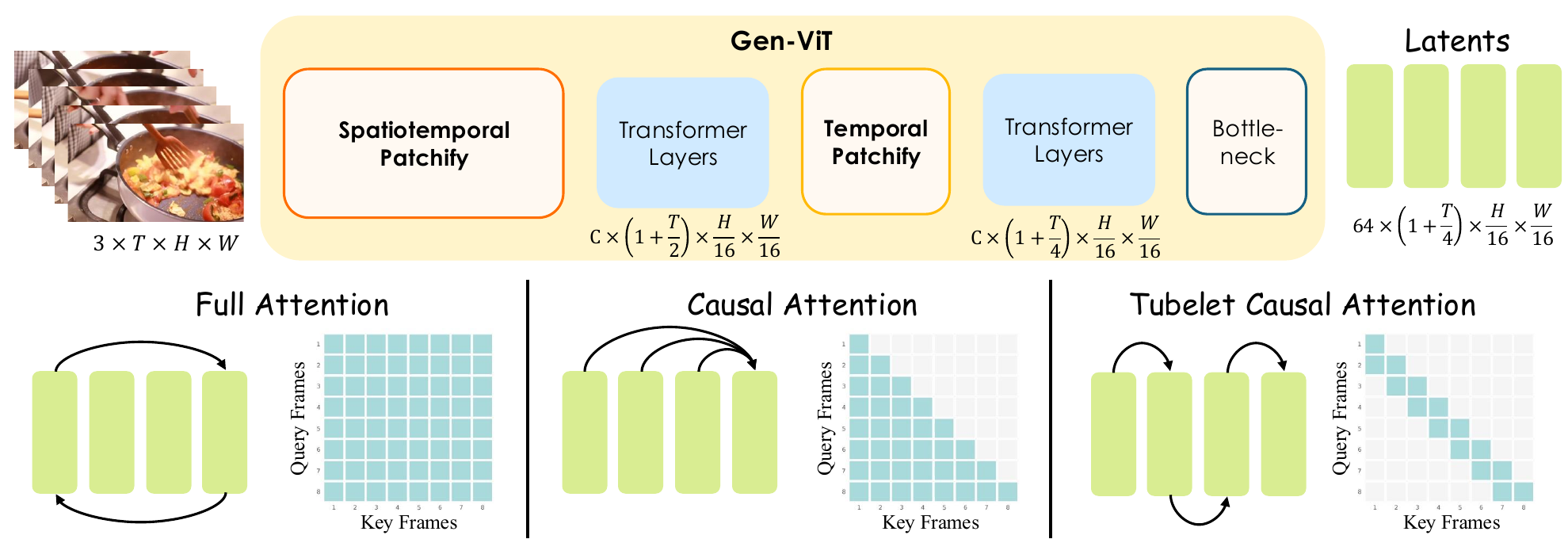}
  \caption{\textbf{Spatiotemporal reconstruction design.} (Top) The Gen-ViT folds a clip into a compact latent. (Bottom) Three ablated attention masks: Full attends across all space-time tokens, Causal masks future frames, and Tubelet restricts attention to a 2-frame window.}
  \label{fig:rec_arch}
\end{figure}
\begin{table}[t]
  \centering
  \caption{\textbf{Reconstruction ablation on ImageNet ($256\!\times\!256$) and DAVIS ($17\!\times\!256\!\times\!256$).} All three attention-mask baselines use the single-step $4{\times}$ temporal patchify; only the `Ours' row uses the hierarchical $2{\times}2$ schedule. Latency is measured per forward pass on a $17\!\times\!512\!\times\!512$ video clip. Further reconstruction comparisons and visualizations are provided in Appendix~\ref{appendix:rec:512} and \ref{appendix:rec:video}}
  \label{tab:rec_main}
  \resizebox{0.8\linewidth}{!}{%
  \begin{tabular}{cc ccc ccc}
    \toprule
    \multirow{2}{*}{\textbf{Method}}
      & \multirow{2}{*}{\textbf{Latency (s)}\,($\downarrow$)}
      & \multicolumn{3}{c}{\textbf{ImageNet}}
      & \multicolumn{3}{c}{\textbf{DAVIS}}\\
    \cmidrule(lr){3-5}\cmidrule(lr){6-8}
      &
      & \textbf{PSNR}\,($\uparrow$) & \textbf{SSIM}\,($\uparrow$) & \textbf{rFID}\,($\downarrow$)
      & \textbf{PSNR}\,($\uparrow$) & \textbf{SSIM}\,($\uparrow$) & \textbf{rFVD}\,($\downarrow$)\\
    \midrule
    Full attention     & 0.49          & 31.10 & 0.8890 & 0.367      & 27.40 & 0.8277 & 16.20\\
    Causal attention   & 0.45          & 31.38 & 0.8901 & 0.352      & 27.62 & 0.8283 & 14.05\\
    Tubelet attention  & \textbf{0.17} & 31.42 & 0.8907 & 0.347      & 27.69 & 0.8287 & 13.69\\
    \rowcolor{gray!10}
    Ours               & 0.25          & \textbf{31.73} & \textbf{0.8936} & \textbf{0.329}
                                       & \textbf{27.97} & \textbf{0.8307} & \textbf{11.19}\\
    \bottomrule
  \end{tabular}}
\end{table}

Existing ViT-based tokenizers that jointly handle images and videos reconstruction, such as AToken~\citep{atoken} and OmniTokenizer~\citep{omnitokenizer}, share two design choices: full spatiotemporal attention across all frames, and a single-step temporal patchify applied at the input that aggressively compresses the temporal axis. Both choices come at a cost. Full spatiotemporal attention scales quadratically with the clip length and tends to disrupt the per-frame structural prior inherited from image pretraining; the aggressive single-step patchify collapses fine-grained temporal details before any cross-frame reasoning. This naturally raises a critical question: are these design choices really necessary?

We answer this through a controlled ablation along the same two axes: \emph{(i)} the temporal attention region, and \emph{(ii)} the temporal patchify schedule. Following the common design in video VAEs, a clip $\mathbf{x}\!\in\!\mathbb{R}^{3\times(1+T)\times H\times W}$ is encoded into an anchor image latent together with the remaining $T$ frames compressed by a factor of $4$, producing a compact latent $\mathbf{z}\!\in\!\mathbb{R}^{C\times(1+\tfrac{T}{4})\times\tfrac{H}{16}\times\tfrac{W}{16}}$. The two axes are then ablated independently. For \emph{(i)} we compare three attention masks (Fig.~\ref{fig:rec_arch}, bottom): \emph{Full attention}, the standard choice in AToken and OmniTokenizer; \emph{Causal attention}, with a causal mask across all preceding frames; and \emph{Tubelet attention}, where causal attention is restricted to a 2-frame tubelet so each token attends only to its own frame and the immediately preceding one. For \emph{(ii)} we compare the single-step $4\times$ temporal patchify used by AToken and OmniTokenizer against a hierarchical schedule that applies two consecutive $2\times$ patchify stages (top of Fig.~\ref{fig:rec_arch}). During each temporal patchify stage, the anchor frame is zero-padded so that it goes through the same operation as the remaining frames.

Table~\ref{tab:rec_main} reveals two principles that contradict the common choices. First, expanding the temporal receptive field beyond a 2-frame tubelet only degrades reconstruction: both full bidirectional and all-past causal attention perform worse than Tubelet attention. Second, distributing temporal compression across two patchify stages consistently outperforms a single-step counterpart at the same compression ratio. These results answer our opening question: aggressive spatiotemporal attention and single-step patchify are not only unnecessary but actively suboptimal.

\begin{keyfindingsbox}
\begin{itemize}[leftmargin=1.5em,itemsep=0.35em,topsep=0.35em]
    \item \textbf{Less attention is more.} Restricting causal attention to a 2-frame tubelet  yields the best reconstruction; wider receptive fields disturb local detail more than they help.

    \item \textbf{Hierarchical patchify outperforms single-step.} Distributing temporal compression across two $2\times$ patchify stages consistently improves reconstruction over a single $4\times$ patchify, indicating that the temporal axis benefits from progressive, multi-scale folding.
\end{itemize}
\end{keyfindingsbox}

\subsection{Spatiotemporal Semantic Distillation via the Decompressor}
\label{sec:tokenizer:und}

Following HYDRA, we inject semantic structure into the latent by distilling the Sem-ViT output against pretrained teachers. Extending this recipe to video, however, reveals a fundamental asymmetry. For images, the Sem-ViT output has the same spatial resolution as a frame and can be aligned token-by-token with an off-the-shelf image teacher. For video, the Sem-ViT output is temporally compressed to $1+T/4$ tokens, while existing video encoders  operate at the original frame rate. The video stream therefore receives no video-level semantic supervision under the standard distillation recipe.

\begin{figure}[t]
  \centering
  \includegraphics[width=0.9\linewidth]{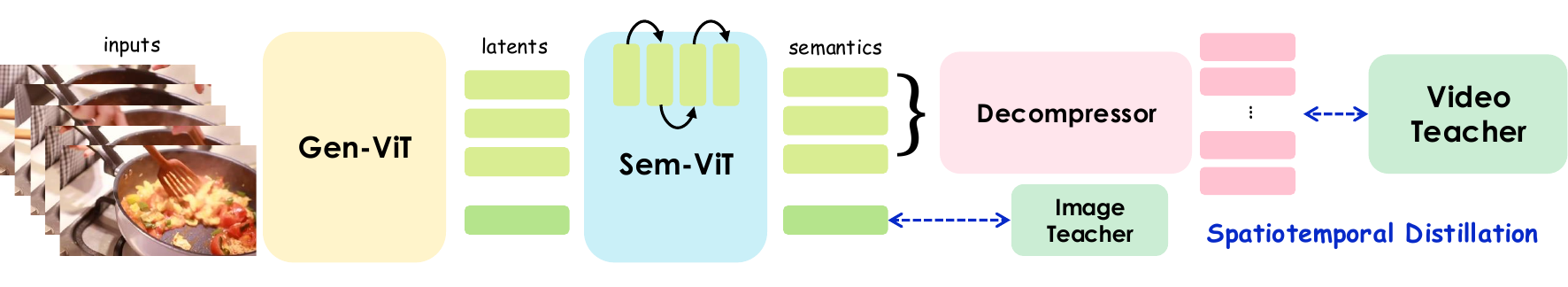}
  \caption{\textbf{Spatiotemporal distillation.} The uncompressed image latent is directly distilled by an image teacher; the $4{\times}$ temporally-compressed video latent is first lifted to origin length $T$ by a lightweight Decompressor before distillation by a video teacher.}
   \vspace{-5mm}
  \label{fig:und_arch}
\end{figure}

\begin{table}[t]
  \centering
  \caption{\textbf{Semantic-distillation ablation.} }
  \label{tab:und_decomp}
  \resizebox{0.8\linewidth}{!}{%
  \begin{tabular}{cccc cc cc c c}
    \toprule
    \multicolumn{4}{c}{\textbf{Design choices}}
      & \multicolumn{2}{c}{\textbf{Vid.\ Und.}}
      & \multicolumn{2}{c}{\textbf{Img.\ Und.}}
      & \textbf{Img.\ Gen.}
      & \textbf{Edit}\\
    \cmidrule(lr){1-4}\cmidrule(lr){5-6}\cmidrule(lr){7-8}\cmidrule(lr){9-9}\cmidrule(lr){10-10}
    \textbf{img} & \textbf{Decomp} & \textbf{Decomp} & \textbf{Sem-ViT}
      & \textbf{MVBench} & \textbf{VideoMME}
      & \textbf{AI2D} & \textbf{MME}
      & \textbf{GenEval}
      & \textbf{ImgEdit}\\
    \textbf{distill} & \textbf{w/ img} & \textbf{w/ video} & \textbf{bi-dir}
      & ($\uparrow$) & ($\uparrow$)
      & ($\uparrow$) & ($\uparrow$)
      & ($\uparrow$)
      & ($\uparrow$)\\
    \midrule
            &           &            &           & 29.8 & 27.4 & 45.1 &  989 & 67.5  & 2.35\\
    \checkmark &           &            &           & 42.1 & 42.5 & 61.2 & 1339 & 70.6  & 2.72\\
    \checkmark & \checkmark &            &           & 44.7 & 44.3 & \textbf{62.7} & \textbf{1522} & 70.7  & 3.07\\
    \rowcolor{gray!10}
    \checkmark &           & \checkmark &           & \textbf{45.4} & \textbf{45.0} & 62.5 & 1501 & \textbf{72.0} & \textbf{3.20}\\
    \checkmark &           & \checkmark & \checkmark & 43.1 & 43.7 & 62.0 & 1434 & 70.1  & 2.70\\
    \bottomrule
  \end{tabular}}
  \vspace{-4mm}
\end{table}

We resolve this asymmetry by introducing a lightweight \emph{Decompressor}, a small ViT module $\mathbf{D}$ that lifts the temporally compressed Sem-ViT output back to its native temporal length, producing dense per-frame semantic features that can be aligned with both image and video teachers (Fig.~\ref{fig:und_arch}). The Decompressor is only used at tokenizer-training time and is discarded afterwards; the LLM still operates on the same compact Sem-ViT output $\mathbf{s}$. Letting $d_{\cos}(\mathbf{a},\mathbf{b})\!=\!1-\cos(\mathbf{a},\mathbf{b})$ denote the cosine distance, the full distillation loss combines an image-teacher term at $\mathbf{s}$ and a video-teacher term at the Decompressor output:
\begin{equation}
  \mathcal{L}_{\text{dist}}
  = d_{\cos}\!\bigl(\mathbf{s}_{0},\,\mathcal{T}_{\text{img}}(\mathbf{x})\bigr)
  + d_{\cos}\!\bigl(\mathbf{D}(\mathbf{s}_{1:}),\,\mathcal{T}_{\text{vid}}(\mathbf{x})\bigr),
  \label{eq:decomp}
\end{equation}

where $\mathbf{s}_{0}$ is the leading uncompressed image token and $\mathbf{s}_{1:}$ are the compressed video latents. For pure image batches, the video term in Eq.~\ref{eq:decomp} is masked out. We ablate four design choices in Table~\ref{tab:und_decomp}: \emph{(i)} whether to apply image distillation at the Sem-ViT output (\emph{img distill}); \emph{(ii)} whether to additionally distill the Decompressor output against an image teacher (\emph{Decomp w/ img}); \emph{(iii)} or against a video teacher (\emph{Decomp w/ video}); and, as a cross-check of F1, \emph{(iv)} whether the Sem-ViT uses bidirectional rather than tubelet attention (\emph{Sem-ViT bi-dir}).

Table~\ref{tab:und_decomp} surfaces three principles. First, semantic distillation is indispensable: removing it collapses both image and video understanding. Second, the Decompressor is what unlocks video-level supervision: distilling it against a video teacher yields the strongest video understanding while preserving image-side performance, and the same configuration also delivers the best image generation and editing scores, consistent with the hypothesis that semantically richer latents accelerate the LLM's convergence on generation and editing. Third, switching the Sem-ViT to bidirectional attention uniformly degrades every metric, mirroring F1: less attention is more even on the understanding side.

\begin{keyfindingsbox}
\begin{itemize}[leftmargin=1.5em,itemsep=0.35em,topsep=0.35em]
    \item \textbf{A semantic latent lifts both understanding and generation.} Dual image- and video-teacher distillation, enabled by the Decompressor, equips the compact latent with explicit spatiotemporal semantic structure, jointly improving understanding and generation.
\end{itemize}
\end{keyfindingsbox}

\section{\modelname: Advancing Unified Multimodal Models with Holistic Tokenizers}
\label{sec:task}

\subsection{Overall Architecture}
\label{sec:task:tpl}

\begin{figure}[t]
  \centering
  \includegraphics[width=1\linewidth]{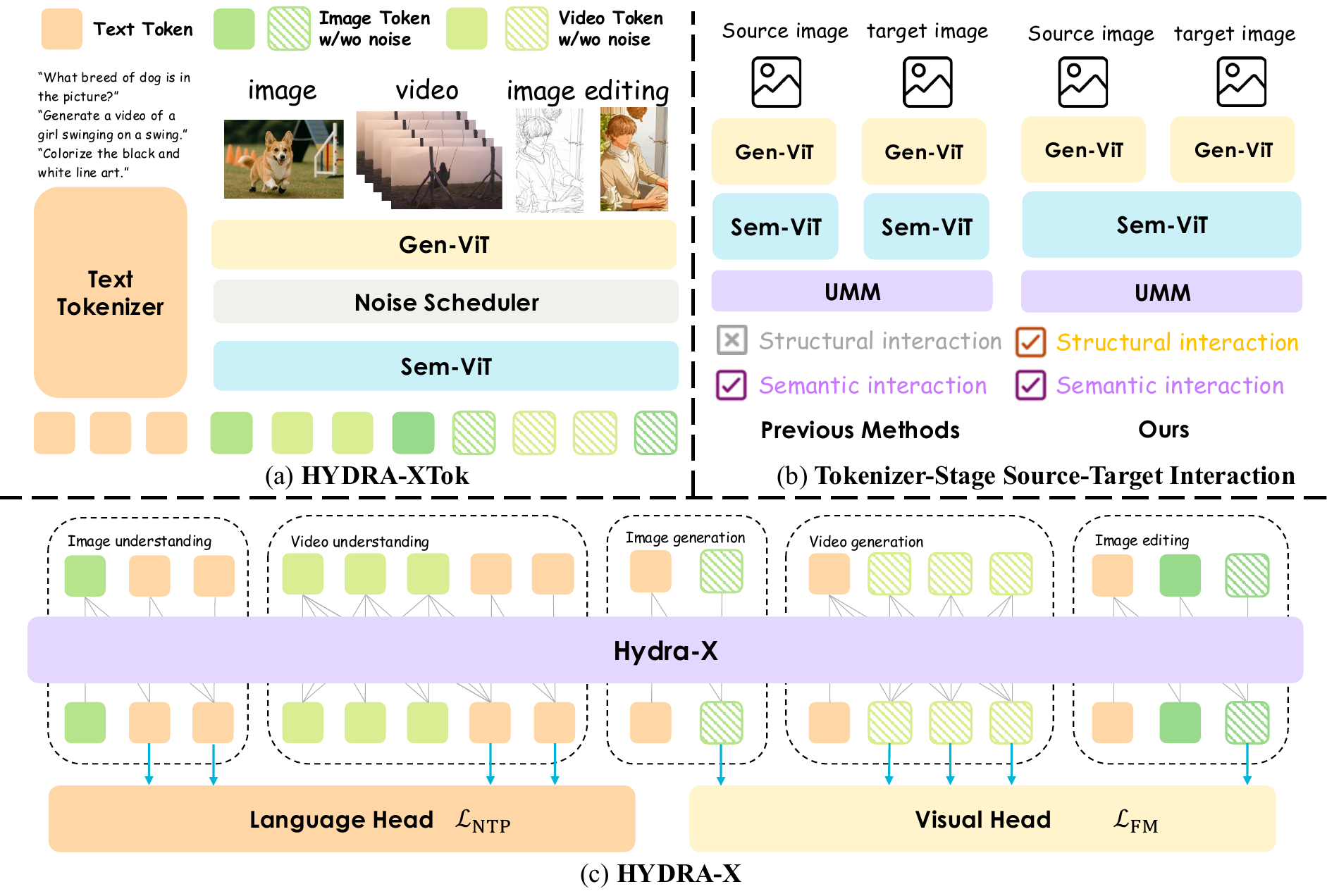}
  \caption{\textbf{\modelname unifies five visual tasks through the holistic tokenizer \tokname.}
\textbf{(a)} \tokname encodes any image or video into a compact Gen-ViT latent  and then into  semantic features with Sem-ViT.
\textbf{(b)} Previous editing pipelines (left) encode source and target with two independent branches; \modelname (right) keeps Gen-ViT independent for faithful reconstruction but shares the Sem-ViT with tubelet causal attention, injecting structural interaction inside the tokenizer.
\textbf{(c)} A shared backbone with two separate heads  drives all five tasks.}
  \label{fig:umm_arch}
\end{figure}

\modelname follows the standard native UMM template~\citep{xie2025show,liu2025tuna,hydra}: text tokens and visual tokens produced by \tokname are interleaved into a single sequence and processed by a shared LLM backbone with two specialised heads, an autoregressive language head trained with next-token prediction and a vision head trained with flow matching~\citep{lipman2022flow,esser2024scaling}. Within this template, \modelname unifies five tasks under one shared tokenizer \tokname (Fig.~\ref{fig:umm_arch}(a)): image generation (text $\to$ image), image understanding (image $\to$ text), video generation (text $\to$ video), video understanding (video $\to$ text), and image editing (source image with text instruction $\to$ target image).

As illustrated in Fig.~\ref{fig:umm_arch}(c), the same Gen-ViT serves all five tasks; the only task-dependent component is which head decode the LLM output. The model is trained end-to-end with the composite loss
\begin{equation}
  \mathcal{L}_{\modelname} \;=\;  \lambda_{\text{1}}\mathcal{L}_{\text{NTP}} + \lambda_{\text{2}}\mathcal{L}_{\text{FM}},
\end{equation}
where $\mathcal{L}_{\text{NTP}}$ is the next-token prediction loss for text, $\mathcal{L}_{\text{FM}}$ is the rectified flow matching loss for visual latents, and both loss weights $\lambda_{1}$ and $\lambda_{2}$ are set to $1$ by default.

\subsection{Independent Encoding Bypasses the Latent}
\label{sec:task:edit}

Among the five tasks, image editing is the only one whose input contains both a conditioning image and a target image. Conventional pipelines, including HYDRA~\citep{hydra} and cascaded designs such as Show-o2~\citep{xie2025show} and TUNA~\citep{liu2025tuna}, tokenise the source $\mathbf{x}_c$ and the target $\mathbf{x}_t$ \emph{independently} with the same tokenizer (Fig.~\ref{fig:umm_arch}(b), left):
\begin{equation}
  [\mathbf{s}_c,\,\mathbf{s}_t]
  \;=\;\bigl[\tokname(\mathbf{x}_c),\,\tokname(\mathbf{x}_t)\bigr],
  \qquad
  \mathbf{s}_c \perp \mathbf{s}_t \quad\text{inside the tokenizer.}
  \label{eq:standard-edit}
\end{equation}
As a result, the source and target latents $\mathbf{z}_c,\mathbf{z}_t$ never interact inside the tokenizer, and the LLM has to discover their cross-image alignment from scratch on top of two independent semantic streams. This is sufficient for high-level semantic edits but consistently fails on detail-faithful edits.

\subsection{Tokenizer-Stage Source-Target Interaction}
\label{sec:task:our}

A natural fix falls out of \tokname's holistic design: since the Sem-ViT already applies tubelet causal attention for video modeling, we reuse the exact same mechanism for editing pairs by routing $(\mathbf{x}_c,\mathbf{x}_t)$ through \tokname as a length-2 clip  (Fig.~\ref{fig:umm_arch}(b), right). The Gen-ViT continues to encode the two images independently and the post-Bottleneck latents $\mathbf{z}_c,\mathbf{z}_t$ remain reconstruction-faithful. The cross-image interaction is then injected exclusively at the Sem-ViT, which processes $[\mathbf{z}_c;\mathbf{z}_t]$ with the same tubelet causal mask used for video:
\begin{equation}
  [\mathbf{s}_c,\,\mathbf{s}_t]
  \;=\;\text{Sem-ViT}\bigl([\mathbf{z}_c;\mathbf{z}_t]\bigr),
  \quad
  \text{causal: } \mathbf{s}_c \text{ attends only to } \mathbf{z}_c, \;\;
  \mathbf{s}_t \text{ attends to } [\mathbf{z}_c;\mathbf{z}_t].
  \label{eq:edit-video}
\end{equation}
Note that for editing pairs we disable Gen-ViT's cross-frame tubelet attention since the source and target are not temporally adjacent video frames; only Sem-ViT (the semantic stage) reuses the video tubelet causal mask. This asymmetric reuse is a deliberate choice: structural reconstruction benefits from independent encoding, while semantic alignment benefits from cross-image interaction.

\begin{table}[t]
  \centering
  \caption{\textbf{Source-target interaction (STI) ablation.} \modelname-STI tokenises the editing pair as a length-2 clip with Sem-ViT tubelet causal attention; \modelname-Indep encodes the source and target independently. \emph{Recon-PSNR}: PSNR of source reconstruction on ImgEdit.}
  \label{tab:edit_abl}
  \resizebox{\linewidth}{!}{%
  \begin{tabular}{l cc c cc cc}
    \toprule
    \multirow{2}{*}{\textbf{Layout}}
      & \multicolumn{2}{c}{\textbf{Editing}}
      & \textbf{Img. Gen.}
      & \multicolumn{2}{c}{\textbf{Vid. Und.}}
      & \multicolumn{2}{c}{\textbf{Img. Und.}}\\
    \cmidrule(lr){2-3}\cmidrule(lr){4-4}\cmidrule(lr){5-6}\cmidrule(lr){7-8}
      & \textbf{ImgEdit}\,($\uparrow$) & \textbf{Recon-PSNR}\,($\uparrow$)
      & \textbf{GenEval}\,($\uparrow$)
      & \textbf{MVBench}\,($\uparrow$) & \textbf{VideoMME}\,($\uparrow$)
      & \textbf{AI2D}\,($\uparrow$) & \textbf{MME}\,($\uparrow$)\\
    \midrule
    \modelname-Indep         & 2.80 & 20.74 & 70.51 & 45.3 & 45.0 & 62.2 & 1478.5\\
    \rowcolor{gray!10}
    \modelname-STI  & \textbf{3.20} & \textbf{27.65} & \textbf{71.97} & \textbf{45.4} & \textbf{45.0} & \textbf{62.5} & \textbf{1501.0}\\
    \bottomrule
  \end{tabular}}
\end{table}

Table~\ref{tab:edit_abl} compares \modelname-Indep against \modelname-STI, identical except for whether the editing pair is encoded independently or as a length-2 clip with Sem-ViT tubelet causal attention. STI raises \emph{Recon-PSNR}, the PSNR of source reconstruction on ImgEdit~\citep{ye2025imgedit} that directly probes editing consistency, by nearly $7$ dB and lifts ImgEdit by $0.4$. STI further yields consistent gains on most non-editing benchmark, with GenEval ($+1.46$) the most prominent, suggesting that the new latent-level coupling also enriches the Sem-ViT for generation. The Recon-PSNR jump directly validates our hypothesis from Section~\ref{sec:task:edit}: editing's consistency failure stems from latent-level isolation inside the tokenizer, not from LLM capacity or supervision.

\begin{keyfindingsbox}
\begin{itemize}[leftmargin=1.5em,itemsep=0.35em,topsep=0.35em]
    \item \textbf{Latent-level interaction matters for editing.} Reusing the Sem-ViT's tubelet causal attention to fuse source and target inside the tokenizer, adds no parameters and no separate cross-image module yet substantially improves editing consistency.
\end{itemize}
\end{keyfindingsbox}

\section{Main Results}
\label{sec:exp}

\paragraph{Implementation.}
\modelname is instantiated at two scales. The reported model uses Qwen2.5\,-7B-Instruct~\citep{qwen2.5} as the LLM backbone; a matched 1.5B variant is used for the methodological ablations in Sections~\ref{sec:tokenizer}--\ref{sec:task}. Following AToken~\citep{atoken}, \tokname includes a symmetric ViT encoder/decoder pair augmented with 3D rotary position embeddings (3D RoPE)~\citep{su2024roformer} for joint spatiotemporal modelling. The Decompressor $\mathbf{D}$ in Eq.~\ref{eq:decomp} is a lightweight $4{\times}$ temporal upsampler that stacks two consecutive \emph{(temporal upsample $\to$ transformer block)} stages; each temporal upsample is a $1{\times}1$ convolution doubling the channel dimension ($C\!\to\!2C$) followed by a channel-to-time reshape, inverting the encoder's hierarchical $2{\times}2$ temporal patchify. The bottleneck dimension is $C\!=\!64$. For distillation teachers, we use SigLIP-SO400M-patch16-naflex~\citep{tschannen2025siglip} as the image teacher $\mathcal{T}_{\text{img}}$ and InternVideo-Next-L~\citep{wang2025internvideo} as the video teacher $\mathcal{T}_{\text{vid}}$.

\subsection{Multimodal Understanding}
\label{sec:exp:und}

\paragraph{Image understanding.}
We benchmark on AI2D~\citep{kembhavi2016ai2d}, MME~\citep{mme}, MMMU~\citep{yue2024mmmu}, OCRBench~\citep{liu2024ocrbench}, MMBench~\citep{mmbench}, RealWorldQA, ChartQA~\citep{masry2022chartqa}, DocVQA~\citep{mathew2021docvqa}, and InfoVQA~\citep{mathew2022infographicvqa}. Table~\ref{tab:img_und} compares \modelname against open-source UMMs at a similar scale. Overall, \modelname matches or exceeds 7B native UMM baselines on most reported metrics, including OCR- and chart-heavy tasks where strong semantic retention is important.

\begin{table*}[!t]
    \centering
    \caption{\textbf{Evaluation on image understanding benchmarks.} \# Params. denotes the model size. Rows in gray indicate models with $\geq 14$B parameters and are excluded from the ranking. Within each subgroup of the  table, \textbf{bold} marks the best result and \underline{underline} marks the second-best.
    }
    \begin{adjustbox}{max width=\linewidth}
        \normalsize
    \begin{tabular}{@{}lc ccccccccc @{}}
    \toprule
    \multirow{2}{*}{\textbf{Models}} & \multirow{2}{*}{\textbf{\# Params}} & \textbf{AI2D} & \textbf{MME} & \textbf{MMMU} & \textbf{OCRBench} & \textbf{MMB} & \textbf{RWQA} & \textbf{ChartQA} & \textbf{DocVQA} & \textbf{InfoVQA} \\
    \cmidrule(lr){3-11}
     & & \textbf{test} & \textbf{summary} & \textbf{val} & \textbf{test} & \textbf{dev\_en} & \textbf{test} & \textbf{test} & \textbf{val} & \textbf{val} \\
    \midrule
    \multicolumn{11}{l}{\textit{Understanding-only Models}} \\
    \midrule
    Qwen2.5-VL \citep{bai2025qwen2} & 7B & \textbf{84.3} & \textbf{2312.0} & \textbf{58.0} & \textbf{88.8} & \textbf{82.8} & \textbf{68.4} & \textbf{84.1} & \textbf{93.0} & \textbf{78.6} \\
    LLaVA-1.5 \citep{liu2023visual} & 7B & 55.5 & 1510.7 & 35.7 & 31.8 & 62.3 & 54.8 & 17.9 & -- & -- \\
    LLaVA-OV \citep{li2024llava} & 7B & \underline{81.4} & \underline{1998.1} & \underline{48.8} & \underline{62.2} & \underline{80.8} & \underline{66.3} & \underline{80.0} & \underline{87.5} & \underline{68.8} \\ 
    \midrule
    \multicolumn{11}{l}{\textit{Unified Multimodal Models}} \\
    \midrule
    BLIP3-o \citep{blip3-o} & 8B & -- & \underline{2329.7} & \underline{50.6} & \underline{83.1} & \underline{83.5} & \underline{69.0} & 78.0 & -- & -- \\
    \g{TokenFlow-XL \citep{qu2025tokenflow}} & \g{14B} & \g{--} & \g{1922.1} & \g{43.2} & \g{--} & \g{68.9} & \g{56.6} & \g{--} & \g{--} & \g{--} \\
    \g{Ming-UniVision \citep{huang2025ming}} & \g{16B} & \g{82.8} & \g{2023.0} & \g{40.3} & \g{72.4} & \g{78.5} & \g{--} & \g{--} & \g{--} & \g{--} \\
    \g{BAGEL \citep{bagel}} & \g{14B} & \g{89.2} & \g{2388.0} & \g{55.3} & \g{73.3} & \g{85.0} & \g{72.8} & \g{78.5} & \g{--} & \g{--} \\
    Janus-Pro \citep{chen2025janus} & 7B & 71.3 & 1567.1 & 41.0 & 59.0 & 79.2 & 58.0 & 25.8 & -- & -- \\
    VILA-U \citep{wu2024vila} & 7B & -- & 1401.8 & -- & -- & 66.6 & -- & -- & -- & -- \\
    Show-o2 \citep{xie2025show} & 7B & 78.6 & 1620.5 & 48.9 & 32.4 & 79.3 & 64.7 & 52.3 & \underline{77.3} & \underline{45.8} \\
    HYDRA \citep{hydra} & 7B & \underline{85.1} & 2068.6 & 49.4 & 57.7 & 82.4 & 64.7 & -- & -- & -- \\ 
    \rowcolor{gray!10} 
    \modelname & 7B & \textbf{86.5} & \textbf{2350.0} & \textbf{51.5} & \textbf{84.5} & \textbf{84.0} & \textbf{68.7} & \textbf{86.5} & \textbf{81.7} & \textbf{59.1} \\ 
    \bottomrule
    \end{tabular}
    \end{adjustbox}
    \label{tab:img_und}
\end{table*}

\paragraph{Video understanding.}
\begin{table}[t]
    \centering
    \caption{\textbf{Evaluation on video understanding benchmarks.} \# Params. denotes the model size. Video-MME reports the w/o-subtitle score. }
    \label{tab:video_und}
    \resizebox{0.75\linewidth}{!}{%
    \begin{tabular}{@{}lc cccc @{}}
    \toprule
    \multirow{2}{*}{\textbf{Models}} & \multirow{2}{*}{\textbf{\# Params}} & \textbf{MVBench} & \textbf{Video-MME} & \textbf{LongVideoBench} & \textbf{LVBench} \\
    \cmidrule(lr){3-6}
     & & \textbf{test} & \textbf{w/o sub} & \textbf{val} & \textbf{test} \\
    \midrule
    \multicolumn{6}{l}{\textit{Proprietary Und.-only Models}} \\
    \midrule
    GPT-4V \citep{openai2023gpt4v}            & --   & \underline{43.5} & 59.9 & 61.3 & --   \\
    GPT-4o \citep{OpenAI_GPT4o}            & --   & --   & \underline{71.9} & \textbf{66.7} & \textbf{48.9} \\
    Gemini-1.5-Flash \citep{team2024gemini}& --   & --   & 70.3 & 61.6 & --   \\
    Gemini-1.5-Pro \citep{team2024gemini}  & --   & \textbf{54.2} & \textbf{75.0} & \underline{64.0} & \underline{33.1} \\
    \midrule
    \multicolumn{6}{l}{\textit{Open-source Und.-only Models}} \\
    \midrule
    VILA \citep{vila}                      & 40B  & --   & \textbf{60.1} & --   & --   \\
    PLLaVA \citep{xu2024pllava}            & 34B  & \underline{58.1} & --   & \underline{53.2} & --   \\
    LongVA \citep{zhang2024long}           & 7B   & 49.2 & 52.6 & 51.8 & --   \\
    VideoLLaMA2 \citep{cheng2024videollama}& 7B   & 54.6 & 47.9 & --   & --   \\
    LLaVA-OV \citep{li2024llava}           & 7B   & 56.7 & \underline{58.2} & \textbf{56.5} & \textbf{26.9} \\
    IXC-2.5 \citep{zhang2024internlm}      & 7B   & \textbf{69.1} & 55.8 & --   & --   \\
    \midrule
    \multicolumn{6}{l}{\textit{Unified Multimodal Models}} \\
    \midrule
    Show-o2 \citep{xie2025show}            & 1.5B & 49.8 & 48.0 & 49.2 & --   \\
    TUNA \citep{liu2025tuna}               & 1.5B & 54.4 & 49.1 & 49.7 & \underline{27.4} \\
    Show-o2 \citep{xie2025show}            & 7B   & \underline{55.8} & \underline{57.4} & \underline{55.5} & --   \\
    \rowcolor{gray!10}
    \modelname                            & 7B   & \textbf{59.1} & \textbf{60.0} & \textbf{59.5} & \textbf{30.0} \\
    \bottomrule
    \end{tabular}}
\end{table}

We evaluate on MVBench~\citep{li2024mvbench}, Video-MME~\citep{fu2025video}, LVBench~\citep{wang2025lvbench}, and LongVideoBench~\citep{wu2024longvideobench}(Table~\ref{tab:video_und}). \modelname improves over the reported 1.5B and 7B unified baselines on the benchmarks where comparable numbers are available. It remains below the strongest dedicated or proprietary video LMMs on several metrics, but narrows the gap while using a single ViT tokenizer shared across understanding, generation, and editing. These results are consistent with the role of dual-teacher distillation in \tokname, which provides the compressed latent with both image- and video-level semantics.

\subsection{Visual Generation}
\label{sec:exp:gen}

Table~\ref{tab:combined_generation} jointly reports image generation on GenEval~\citep{ghosh2023geneval} and WISE~\citep{niu2025wise}, and video generation on VBench~\citep{huang2024vbench} for $17$-frame outputs at $640\times 384$, summarised by Quality Score (QS), Semantic Score (SS), and the aggregate Total score. Among 7B-scale unified baselines, \modelname is the strongest entry on every reported GenEval and WISE column; compared against $\geq14$B unified models, it remains competitive on the Overall scores while using a 7B backbone. On VBench, \modelname leads all unified entries on QS, SS, and Total, improving over the closest unified competitor (Show-o2-1.5B) by $+1.87$ QS, $+3.26$ SS, and $+2.15$ Total. Per-dimension VBench scores are provided in Appendix Table~\ref{tab:vbench}, where \modelname additionally leads in semantic-heavy dimensions including Object Class, Human Action, and Scene. Together these results suggest that dual-teacher distillation transfers semantic structure into the latent while preserving its role in visual synthesis.

\begin{table*}[t]
    \centering
    \caption{\textbf{Comprehensive visual generation results.} Image generation on {GenEval}~\citep{ghosh2023geneval} and {WISE}~\citep{niu2025wise}; video generation on {VBench}~\citep{huang2024vbench} reporting Quality Score (\textbf{QS}), Semantic Score (\textbf{SS}), and the aggregate \textbf{Total} score. $^\dagger$ refers to using LLM rewriters. Rows in gray indicate models with $\geq 14$B parameters and are excluded from the ranking. Qualitative results are in Appendix~\ref{appendix_t2i}.}
    \label{tab:combined_generation}
    \begin{adjustbox}{max width=\linewidth}
    \scriptsize
    \begin{tabular}{@{}lc ccc c ccc c ccc @{}}
    \toprule
    \multirow{2}{*}{\textbf{Models}} & \multirow{2}{*}{\textbf{\# Params}} & \multicolumn{3}{c}{\textbf{GenEval}} & & \multicolumn{3}{c}{\textbf{WISE}} & & \multicolumn{3}{c}{\textbf{VBench}} \\
    \cmidrule(lr){3-5} \cmidrule(lr){7-9} \cmidrule(lr){11-13}
     & & \textbf{Two Obj.} & \textbf{Pos.} & \textbf{Over.} & & \textbf{Cult.} & \textbf{Space} & \textbf{Over.} & & \textbf{QS} & \textbf{SS} & \textbf{Total} \\
    \midrule
    \multicolumn{13}{l}{\textit{Generation-only Models}} \\
    \midrule
    SD3-Med \citep{esser2024scaling}             & 2B   & \textbf{0.94}     & \underline{0.33}  & \underline{0.74}  & & --                & --                & --                & & --                & --                & --                \\
    FLUX.1 [Dev] $^\dagger$ \citep{flux}                    & 12B  & \underline{0.93}  & \textbf{0.68}     & \textbf{0.82}     & & \textbf{0.48}     & \textbf{0.62}     & \textbf{0.50}     & & --                & --                & --                \\
    CogVideoX \citep{yang2024cogvideox}          & 5B   & --                & --                & --                & & --                & --                & --                & & \underline{82.75} & \textbf{77.04}    & \underline{81.61} \\
    Hunyuan Video \citep{kong2024hunyuanvideo}   & 13B  & --                & --                & --                & & --                & --                & --                & & \textbf{85.07}    & \underline{76.88} & \textbf{83.43}    \\
    \midrule
    \multicolumn{13}{l}{\textit{Unified Multimodal Models}} \\
    \midrule
    \g{TokenFlow-XL \citep{qu2025tokenflow}}     & \g{14B}  & \g{0.60}    & \g{0.16}    & \g{0.55}    & & \g{--}    & \g{--}    & \g{--}    & & \g{--}            & \g{--}            & \g{--}            \\
    \g{SEED-X \citep{seed-x}}                    & \g{17B}  & \g{0.58}    & \g{0.19}    & \g{0.49}    & & \g{--}    & \g{--}    & \g{--}    & & \g{--}            & \g{--}            & \g{--}            \\
    \g{Ming-UniVision \citep{huang2025ming}}     & \g{16B}  & \g{0.93}    & \g{0.92}    & \g{0.85}    & & \g{--}    & \g{--}    & \g{--}    & & \g{--}            & \g{--}            & \g{--}            \\
    \g{BAGEL $^\dagger$ \citep{bagel}}                      & \g{14B}  & \g{0.95}    & \g{0.78}    & \g{0.88}    & & \g{0.44}  & \g{0.68}  & \g{0.52}  & & \g{--}            & \g{--}            & \g{--}            \\
    Show-o2 \citep{xie2025show}                  & 1.5B & 0.86              & 0.46              & 0.73              & & 0.33              & 0.53              & 0.39              & & \underline{82.10} & \underline{78.31} & \underline{81.34} \\
    Harmon \citep{wu2025harmonizing}             & 1.5B & 0.86              & 0.74              & 0.76              & & 0.38              & 0.52              & 0.41              & & --                & --                & --                \\
    Blip3-o$^{\dagger}$ \citep{blip3-o}          & 8B   & --                & --                & \underline{0.84}  & & --                & --                & \underline{0.50}  & & --                & --                & --                \\
    
    MUSE-VL \citep{muse-vl}                      & 7B   & 0.64              & 0.25              & 0.57              & & --                & --                & --                & & --                & --                & --                \\
    Janus-Pro \citep{chen2025janus}              & 7B   & \underline{0.89}  & \underline{0.79}  & 0.80              & & 0.30              & 0.49              & 0.35              & & --                & --                & --                \\
    VILA-U \citep{wu2024vila}                    & 7B   & --                & --                & --                & & 0.26              & 0.37              & 0.31              & & 76.26             & 65.04             & 74.01             \\
    HaploOmni \citep{xiao2025haploomni}          & 7B   & --                & --                & --                & & --                & --                & --                & & --                & --                & 78.10             \\
    Emu3 \citep{emu3}                            & 8B   & --                & --                & 0.66              & & --                & --                & --                & & --                & --                & 80.96             \\
    
    Show-o2 $^\dagger$ \citep{xie2025show}                  & 7B   & 0.87              & 0.52              & 0.76              & & \underline{0.40}  & \underline{0.58}  & 0.44              & & --                & --                & --                \\
    \rowcolor{gray!10}
    \modelname                                  & 7B   & \textbf{0.95}     & \textbf{0.84}     & \textbf{0.88}     & & \textbf{0.53}     & \textbf{0.72}     & \textbf{0.56}     & & \textbf{83.97}    & \textbf{81.57}    & \textbf{83.49}    \\
    \bottomrule
    \end{tabular}
    \end{adjustbox}
\end{table*}

\subsection{Image Editing}
\label{sec:exp:edit}

\begin{table}[t]
    \centering
    \caption{\textbf{Image editing.} ImgEdit-Bench: \textbf{Ext.}=Extract, \textbf{Rm.}=Remove, \textbf{Over.}=overall (mean of 9 categories). GEdit-Bench: \textbf{G-SC}=G-Semantic Consistency, \textbf{G-PQ}=G-Perceptual Quality, \textbf{G-Over.}=overall. Per-dimension breakdown is provided in Appendix~\ref{tab:img_edit_detailed}.}
    \label{tab:img_edit}
    \resizebox{0.78\linewidth}{!}{%
    \scriptsize
    \begin{tabular}{@{}l ccc c ccc@{}}
    \toprule
    \multirow{2}{*}{\textbf{Models}} & \multicolumn{3}{c}{\textbf{ImgEdit-Bench}} & & \multicolumn{3}{c}{\textbf{GEdit-Bench}} \\
    \cmidrule(lr){2-4} \cmidrule(lr){6-8}
     & \textbf{Ext.} & \textbf{Rm.} & \textbf{Over.} & & \textbf{G-SC} & \textbf{G-PQ} & \textbf{G-Over.} \\
    \midrule
    \multicolumn{8}{l}{\textit{Generation-only Models}} \\
    \midrule
    FLUX.1 Kontext [Pro] \citep{flux}     & \underline{2.35} & \underline{3.57} & \underline{4.00} & & \underline{7.02} & \underline{7.60} & \underline{6.56} \\
    Qwen-Image \citep{wu2025qwen-image}   & \textbf{3.43}    & \textbf{4.14}    & \textbf{4.27}    & & \textbf{8.00}    & \textbf{7.86}    & \textbf{7.56}    \\
    \midrule
    \multicolumn{8}{l}{\textit{Unified Multimodal Models}} \\
    \midrule
    OmniGen \citep{xiao2025omnigen}       & 1.71             & 2.43             & 2.96             & & 5.96             & 5.89             & 5.06             \\
    UniWorld-V1 \citep{lin2025uniworld}   & \underline{2.27} & \underline{3.24} & 3.26             & & 4.93             & \textbf{7.43}    & 4.85             \\
    \g{BAGEL \citep{bagel}}               & \g{1.70}         & \g{2.62}         & \g{3.20}         & & \g{7.36}         & \g{6.83}         & \g{6.52}         \\
    OmniGen2 \citep{wu2025omnigen2}       & 1.77             & 3.20             & \underline{3.44} & & \underline{7.16} & 6.77             & \underline{6.41} \\
    \rowcolor{gray!10}
    \modelname                            & \textbf{4.04}    & \textbf{4.38}    & \textbf{4.34}    & & \textbf{7.80}    & \underline{7.24} & \textbf{7.17}    \\
    \bottomrule
    \end{tabular}}
\end{table}

Table~\ref{tab:img_edit} reports editing on GEdit-Bench~\citep{gedit} and ImgEdit-Bench~\citep{ye2025imgedit}. Among 7B-scale unified models, \modelname leads on \textbf{Ext.}\,($4.04$, $+1.77$), \textbf{Rm.}\,($4.38$, $+1.14$), ImgEdit \textbf{Over.}\,($4.34$, $+0.90$), and GEdit \textbf{G-SC}/\textbf{G-Over.}\,($7.80$/$7.17$), also beating BAGEL-14B on every column. The largest gains land on \textbf{Ext.}\ and \textbf{Rm.}---both needing identity-faithful source preservation---validating the tokenizer-stage source--target interaction in Section~\ref{sec:task:our}. With a 7B backbone, \modelname trails Qwen-Image-20B by only $0.20$/$0.39$ on \textbf{G-SC}/\textbf{G-Over.}; per-dimension scores in Appendix Table~\ref{tab:img_edit_detailed}.

\section{Conclusion}
\label{sec:conclusion}

We presented \modelname, the first native UMM framework that unifies image and video tokenization within a single ViT. Three counter-intuitive design choices in \tokname, frame-level causal tubelet attention, hierarchical temporal patchify, and a Decompressor for dual image-video teacher supervision, efficiently transform an image tokenizer into a video-and-image tokenizer. Rather than treating image editing as a purely LLM-side problem, we elegantly repurpose our video temporal-causal mechanism to process source and target images as length-2 clips. This restores the fine-grained latent-level coupling that is fundamentally lost in prior independent-encoding pipelines. Through this unified design, the visual tokenizer transcends its traditional role as a static image encoder, emerging as a holistic image-and-video interface that unifies five tasks under one shared backbone.

\bibliography{reference}
\bibliographystyle{iclr2025_conference}

\newpage
\appendix
\section{Training Details}
\label{appendix:training}

\subsection{Tokenizer Training Loss}
\label{tokenizer_loss}
\tokname is designed as the visual interface of \modelname: before any token reaches the LLM, it must be compact enough for generation, faithful enough for reconstruction, and semantic enough for understanding. We initialize Gen-ViT and Sem-ViT from SigLIP\,2~\citep{tschannen2025siglip}. The tokenizer is trained with a reconstruction term and a semantic distillation term:
\begin{equation*}
  \mathcal{L}_{\tokname}
  = \mathcal{L}_{\text{rec}}
  + \lambda_{\text{dist}}\mathcal{L}_{\text{dist}},
\end{equation*}
where $\mathcal{L}_{\text{rec}}$ is the reconstruction term detailed below and $\mathcal{L}_{\text{dist}}$ aligns Sem-ViT features with the image and video teachers (Eq.~\ref{eq:decomp}).

To keep the compact latent both pixel-faithful and structurally stable, the reconstruction term $\mathcal{L}_{\text{rec}}$ encapsulates pixel-level recovery, perceptual fidelity, and latent space regularization. Specifically, it combines an L1 loss for direct pixel-space reconstruction, an LPIPS perceptual loss $\mathcal{L}_{\text{lpips}}$, an adversarial GAN loss $\mathcal{L}_{\text{gan}}$ to refine texture realism, and a Kullback--Leibler (KL) divergence penalty that aligns the posterior with a standard normal prior. The comprehensive reconstruction objective is formulated as:
\begin{equation}
    \mathcal{L}_{\text{rec}} = \lambda_{1} \| \mathbf{x} - \hat{\mathbf{x}} \|_1 + \lambda_{\text{perc}}\mathcal{L}_{\text{lpips}} + \lambda_{\text{gan}}\mathcal{L}_{\text{gan}} - \lambda_{\text{KL}} \sum_{j=1}^{C} \left( 1 + \boldsymbol{\rho}_j - \boldsymbol{\mu}_j^2 - \exp(\boldsymbol{\rho}_j) \right),
\end{equation}
where $\mathbf{x}$ and $\hat{\mathbf{x}}$ are the original and reconstructed images, while $\boldsymbol{\mu}_j$ and $\boldsymbol{\rho}_j$ are the mean and log-variance of the compressed latent. 

\subsection{Tokenizer Pre-training}
\label{tokenizer_training}
\tokname is trained in three progressive stages to balance foundational representation learning with high-fidelity generative quality:

\paragraph{Stage 1: Foundation Training.}
Initialized with SigLIP-2, \tokname first undergoes training on ImageNet-1.2M at $256 \times 256$ resolution. We then transition to mixed-resolution training, combining $256 \times 256$ videos with images ranging from $256$ to $2048$ pixels. This strategy empowers the tokenizer to generalize effectively to high-resolution video. We optimize the model for $300$k iterations using AdamW with a peak learning rate of $2 \times 10^{-4}$, employing a hybrid SigLIP-2 / InternVideo teacher for distillation.

\paragraph{Stage 2: Decoder Refinement.}
To enhance texture realism and perceptual fidelity, we freeze the encoder and exclusively fine-tune the $27$-layer ViT decoder. Adversarial training (GAN loss) is incorporated in this stage to significantly improve visual reconstruction.

\paragraph{Stage 3: Representation Harmonization.}
In the final stage, we first compute the channel-wise mean and standard deviation of the Gen-ViT latent features. We then freeze Gen-ViT and the decoder while unfreezing Sem-ViT. The Gen-ViT features are normalized before being fed into Sem-ViT and the decoder; during this process, only Sem-ViT is updated. This normalization eliminates feature heterogeneity between the two heads and establishes a unified, semantic-aware latent space capable of faithful reconstruction, which is crucial for downstream UMM tasks.

\subsection{Native Unified Multimodal Models Pre-training}
\label{umm_pretraining}
\begin{table}[htbp]
    \centering
    \caption{\textbf{Training details and computational cost of our \modelname.}  The \tokname pre-training takes an additional 24h on 256h GPUs.  \textsuperscript{$\dagger$}Data Ratio denotes Text: Image Caption : Image Generation : Video Caption: Video SFT: Image SFT: Edit.}
    \begin{adjustbox}{max width=0.95\linewidth} 
\begin{tabular}{cccc}
\toprule
Setting & Stage 1 & Stage 2 &  Stage 3 \\
\midrule
 & Vision Head: $10^{-4}$ & Vision Head: $ 5\times10^{-5}$ & Vision Head: $ 5\times10^{-5}$ \\
\multirow{-2}[0]{*}{LR.} & Sem-ViT: $ 5\times10^{-5}$  & LLM \& Sem-ViT: $ 2\times10^{-5}$ & LLM \& Sem-ViT: $ 2\times10^{-5}$ \\
Base Resolution & 256 & 512 & 1024 \\
Batch Size & 1024 & 1024 & 1024 \\
Tasks & Image (Und. \& Gen.) & + Video \& Edit & + Text Und. \\
Data Ratio\textsuperscript{$\dagger$} & 0:1:3:0:0:0:0 & 0:2:6:1:1:0:0 & 1:0:3:3:0:1:3  \\
\midrule
Hardware & 256 GPUs & 512 GPUs & 512 GPUs \\
Training Step & 50K & 200K & 20K \\
Time Cost & $\sim$10h & $\sim$96h & $\sim$24h \\
\bottomrule
\end{tabular}
\end{adjustbox}
    \label{tab:training_details}
\end{table}
To cultivate the harmonized nature of \modelname, we implement a three-stage progressive training strategy for the unified multimodal model. Detailed configurations and computational cost are summarised in Table~\ref{tab:training_details}.

\paragraph{Stage 1: Unified Representation Alignment.}
To resolve the representation divergence at the input level, we freeze the LLM (Qwen2.5-7B-Instruct) and exclusively tune the vision components (projector, time-step embedding, and flow head). Utilizing $100$M image--text pairs, this phase aligns the visual latent space with the linguistic domain, ensuring a coherent unified input representation.

\paragraph{Stage 2: Comprehensive Multimodal Pre-training.}
We unlock all parameters to facilitate harmonized co-promotion within a single unified stream. The model is jointly optimized on a balanced mix of $30$M understanding samples and $30$M generative samples (strategically filtered from Stage~1). We further incorporate approximately $2$M image editing samples and $10$M video samples into the joint training process. This full-parameter update ensures the compatibility of the learning process and allows the diverse tasks to mutually reinforce each other.

\paragraph{Stage 3: High-Quality Instruction Fine-tuning.}
The final stage focuses on high-fidelity refinement using curated datasets. For multimodal understanding (MMU), we employ $6$M instruction-tuning samples sourced from LLaVA-OneVision~\citep{li2024llava} and Pixmo~\citep{pixmo}, alongside $1.2$M video instruction-tuning samples from LLaVA-Video~\citep{zhang2025llavavideo}. For generation, we utilize $10$M aesthetic-filtered images (derived from Stage~2) and $6$M high-fidelity synthetic images. Additionally, we continue to train on high-quality image editing data to further enhance the model's precise control capabilities.

\subsection{Ablation Study Training Details}
\label{subsec:training_data_abla}

In our ablation studies, the evaluation covers three core capabilities with specific setups:
(i) \textbf{Multimodal Understanding:} We train \modelname using the LLaVA-1.5 multimodal understanding dataset~\citep{llava1.5} combined with the LLaVA-Video SFT dataset~\citep{zhang2025llavavideo}.
(ii) \textbf{Image Generation:} We use Qwen2.5-1.5B~\citep{qwen2.5} as the base model, first training on $20$M image-caption pairs, then further fine-tuning with the ImgEdit dataset for image editing capabilities.
(iii) \textbf{Image Reconstruction:} We train on the ImageNet-1k (1.2M) dataset~\citep{imagenet} for $150$k iterations and assess quality using rFID.

\section{Visual Reconstruction}
\label{appendix:rec_quant}
\begin{table*}[!t]
    \centering
    \caption{\textbf{Reconstruction comparison on ImageNet, DAVIS, and UCF.}
    All methods are evaluated with a unified protocol using their official implementations: inputs are resized and center-cropped to $256{\times}256$ and metrics are computed with identical scripts. Compression ratios are reported separately along the spatial ($f_s$) and temporal ($f_t$) axes; image-only tokenizers have $f_t{=}1$. Within each subgroup, \textbf{bold} marks the best result and \underline{underline} marks the second-best. $^{\dagger}$ indicates models trained strictly on the ImageNet-1.2M dataset.
    }
    \vspace{-2mm}
    \begin{adjustbox}{max width=\linewidth}
    \normalsize
    \begin{tabular}{@{}l cc ccc ccc ccc @{}}
    \toprule
    \multirow{2}{*}{\textbf{Method}} & \multicolumn{2}{c}{\textbf{Compression}} & \multicolumn{3}{c}{\textbf{ImageNet}} & \multicolumn{3}{c}{\textbf{DAVIS}} & \multicolumn{3}{c}{\textbf{UCF}} \\
    \cmidrule(lr){2-3} \cmidrule(lr){4-6} \cmidrule(lr){7-9} \cmidrule(lr){10-12}
     & \textbf{Spatial} & \textbf{Temporal} & \textbf{PSNR}\,($\uparrow$) & \textbf{SSIM}\,($\uparrow$) & \textbf{rFID}\,($\downarrow$) & \textbf{PSNR}\,($\uparrow$) & \textbf{SSIM}\,($\uparrow$) & \textbf{rFVD}\,($\downarrow$) & \textbf{PSNR}\,($\uparrow$) & \textbf{SSIM}\,($\uparrow$) & \textbf{rFVD}\,($\downarrow$) \\
    \midrule
    \multicolumn{12}{l}{\textit{Generation-only Tokenizers}} \\
    \midrule
    SD-VAE \citep{sd}                       & $8\times$  & $1\times$ & 26.26             & 0.745             & 0.606             & --                & --                & --                & --                & --                & --                \\
    RAE$^{\dagger}$ \citep{rae}             & $16\times$ & $1\times$ & 18.05             & 0.500             & 2.040             & --                & --                & --                & --                & --                & --                \\
    FLUX.1 [dev] \citep{flux}               & $8\times$  & $1\times$ & \textbf{32.86}    & \textbf{0.917}    & \textbf{0.176}    & --                & --                & --                & --                & --                & --                \\
    Qwen-Image \citep{wu2025qwen-image}     & $8\times$  & $1\times$ & \underline{32.18} & \underline{0.899} & 1.459             & --                & --                & --                & --                & --                & --                \\
    VAVAE$^{\dagger}$ \citep{vavae}         & $16\times$ & $1\times$ & 27.70             & 0.798             & \underline{0.279} & --                & --                & --                & --                & --                & --                \\
    Wan2.2 \citep{wan2.2}                   & $16\times$ & $4\times$ & 31.25             & 0.878             & 0.749             & \textbf{27.64}    & \textbf{0.820}    & \textbf{14.78}    & \textbf{36.11}    & \textbf{0.961}    & \textbf{4.15}     \\
    \midrule
    \multicolumn{12}{l}{\textit{Unified Tokenizers}} \\
    \midrule
    OmniTokenizer \citep{omnitokenizer}     & $8\times$  & $4\times$ & 26.74             & 0.824             & 1.023             & 24.30             & 0.737             & 113.56            & 29.20             & 0.931             & 38.15             \\
    Vila-U \citep{wu2024vila}               & $16\times$ & $1\times$ & 22.24             & 0.612             & 4.231             & --                & --                & --                & --                & --                & --                \\
    UniTok \citep{unitok}             & $16\times$ & $1\times$ & 25.34             & 0.742             & 0.362             & --                & --                & --                & --                & --                & --                \\
    AToken-So/C (Stage 3) \citep{atoken}    & $16\times$ & $4\times$ & 29.72             & 0.848             & \underline{0.209} & \underline{26.60} & \underline{0.784} & \underline{29.19} & \underline{34.66} & \underline{0.953} & \underline{7.77}  \\
    \rowcolor{gray!10}
    \tokname$^{\dagger}$                   & $16\times$ & $1\times$ & \textbf{32.96}    & \textbf{0.905}    & \textbf{0.154}    & --                & --                & --                & --                & --                & --                \\
    \rowcolor{gray!10}
    \tokname (Stage 3)                     & $16\times$ & $4\times$ & \underline{32.04} & \underline{0.898} & 0.465             & \textbf{28.19}    & \textbf{0.835}    & \textbf{11.61}    & \textbf{36.88}    & \textbf{0.967}    & \textbf{3.11}     \\
    \bottomrule
    \end{tabular}
    \end{adjustbox}
    \label{tab:main_rec}
\end{table*}

We benchmark \tokname under a unified protocol against three families of tokenizers: image-only generative VAEs, video VAEs, and joint image--video tokenizers. All inputs are resized and centre-cropped to $256{\times}256$ and metrics are computed with identical scripts. Table~\ref{tab:main_rec} reports PSNR, SSIM, and rFID/rFVD on ImageNet, DAVIS, and UCF.

To isolate the effect of architecture from that of training data, we additionally report \tokname$^{\dagger}$, a controlled variant trained strictly on ImageNet-1.2M to match the data budget of RAE$^{\dagger}$ and VAVAE$^{\dagger}$. Under this matched-data setting, \tokname$^{\dagger}$ outperforms RAE$^{\dagger}$ and VAVAE$^{\dagger}$ by a large margin on every ImageNet metric (e.g., $+5.26$\,dB PSNR over VAVAE$^{\dagger}$). More strikingly, despite operating at \emph{twice} the compression ratio of dedicated $8\times$ image VAEs and using strictly less training data, \tokname$^{\dagger}$ still exceeds the $8\times$ image VAE, FLUX.1, on ImageNet PSNR ($32.96$ vs.\ $32.86$) and rFID ($0.154$ vs.\ $0.176$), indicating that the holistic ViT design rather than the data scale drives the gain. The fully trained \tokname is the strongest unified tokenizer on \emph{every} video metric: it improves over the previous best AToken-So/C by $+1.59$\,dB DAVIS PSNR and $+2.22$\,dB UCF PSNR, while more than halving rFVD on both datasets ($11.61$ vs.\ $29.19$ on DAVIS; $3.11$ vs.\ $7.77$ on UCF). On video benchmarks, \tokname also outperforms the dedicated $16\times$ video VAE Wan~2.2 ($+0.55$\,dB DAVIS PSNR, $+0.77$\,dB UCF PSNR; rFVD reduced by $21\%$ and $25\%$ respectively), suggesting that a single holistic ViT with hierarchical patchify is a competitive alternative to cascaded image$+$video designs.

\section{Evaluation Details of Multi-modal Understanding Benchmarks}
\label{appen:subsec:metric}

To comprehensively evaluate the perception and reasoning capabilities of \modelname, we employ nine diverse benchmarks covering general understanding, expert knowledge, document/chart comprehension, and fine-grained visual perception. We benchmark on AI2D~\citep{kembhavi2016ai2d} (\texttt{test} split), MME~\citep{mme} (\texttt{test} split), MMMU~\citep{yue2024mmmu} (\texttt{val} split), OCRBench~\citep{liu2024ocrbench} (\texttt{test} split), MMBench~\citep{mmbench} (\texttt{dev\_en} split), RealWorldQA (\texttt{test} split), ChartQA~\citep{masry2022chartqa} (\texttt{test} split), DocVQA~\citep{mathew2021docvqa} (\texttt{val} split), and InfoVQA~\citep{mathew2022infographicvqa} (\texttt{val} split). Table~\ref{tab:img_und} compares \modelname against open-source UMMs at a similar scale. Overall, \modelname matches or exceeds 7B native UMM baselines on most reported metrics, including OCR- and chart-heavy tasks that simultaneously require fine-grained visual details (e.g., character strokes, table cells) and rich semantic structure (e.g., layout and relational reasoning), both of which \tokname's compact latent is designed to preserve.

\section{Tokenizer-Stage Source--Target Interaction: Visual Evidence}
\label{appendix:sti}

\begin{figure}[h]
    \centering
    \includegraphics[width=0.92\linewidth]{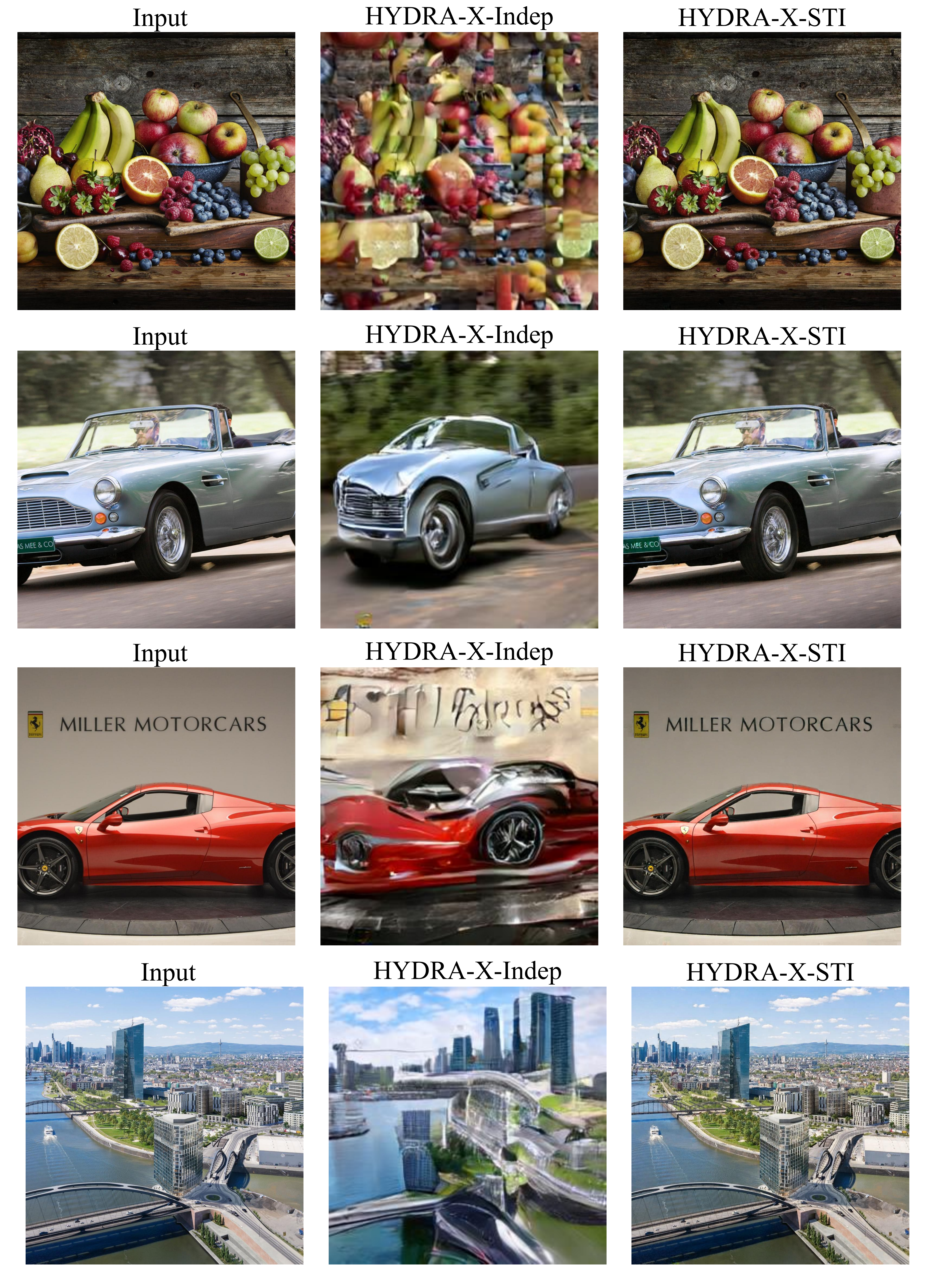}
    \caption{\textbf{Qualitative effect of tokenizer-stage source--target interaction.} Source-image reconstruction produced by \modelname-Indep (independent Sem-ViT encoding of source and target, the conventional pipeline) versus \modelname-STI (joint encoding through tubelet causal attention, our proposal). The two variants share every other architectural component. \modelname-STI preserves identity-sensitive details (object layout, characters, on-screen text) that \modelname-Indep loses, despite both pipelines using the same LLM and the same number of parameters.}
    \label{fig:edit_mechanism}
\end{figure}

Section~\ref{sec:task:our} argues that tokenizer-stage source--target interaction (STI)---routing the source $\mathbf{x}_c$ and target $\mathbf{x}_t$ jointly through a shared Sem-ViT with tubelet causal attention rather than encoding them independently---is the missing ingredient for identity-faithful image editing. Quantitatively, this single change recovers nearly $7$\,dB of source-reconstruction PSNR while leaving the rest of the architecture and parameter count untouched (Table~\ref{tab:edit_abl}). Figure~\ref{fig:edit_mechanism} provides the qualitative counterpart.

We compare two variants of \modelname that differ \emph{only} in this routing step: \textbf{\modelname-Indep} encodes the source and target through two independent Sem-ViT branches---the conventional pipeline shared by BAGEL, OmniGen2, and similar systems---while \textbf{\modelname-STI} routes the pair through a single shared Sem-ViT with tubelet causal attention, treating $(\mathbf{x}_c,\mathbf{x}_t)$ as a length-$2$ clip. Both variants share the same Gen-ViT, the same LLM, and the same flow-matching head, with identical parameter count and training schedule.

The contrast is striking. In the still-life example (top row), \modelname-Indep collapses into a fragmented mosaic where fruit positions, textures, and lighting are all hallucinated locally, whereas \modelname-STI returns a near-pixel-perfect reproduction. The car example (bottom row) makes the failure mode of independent encoding explicit: the Indep variant ``re-imagines'' the car as a different vehicle, removing the driver and passenger and erasing the on-screen text; the STI variant preserves the entire scene including the people inside and the visible plate. These observations confirm the mechanism analysed in Section~\ref{sec:task:our}: in the conventional pipeline, the latent already loses identity-sensitive information \emph{before} the LLM ever reads it, so even a perfectly reasoning LLM cannot recover the source faithfully. STI fixes this bottleneck inside the tokenizer at zero parameter cost, which is precisely what enables the consistent margin on identity-sensitive editing dimensions reported in Tables~\ref{tab:img_edit} and~\ref{tab:img_edit_detailed}.

\section{Limitations}
\label{appendix:limitations}
First, the current scale of training data and model parameters remains a bottleneck, potentially limiting the model's ability to capture the full complexity of high-dimensional video distributions. Second, resource constraints prevented us from exploring long video generation and video editing, both of which are natural extensions of our holistic encoder. Finally, for a fair comparison, we instantiate \modelname only on a 7B \emph{dense} LLM; pairing our tokenizer with more advanced backbones, such as MoE~\citep{li2024aria} or MoT~\citep{liang2024mixture}, represents a clear path to further amplify cross-task performance gains.

\section{Broader Impacts}
\label{appendix:broader}
As \modelname introduces strong text-to-image generation capabilities within a unified framework, we acknowledge potential downstream risks. These include the generation of misleading or fabricated visual content (e.g., deepfakes), which could be exploited for disinformation or impersonation. To mitigate such risks, we advocate for the incorporation of content watermarking and provenance tracking mechanisms upon deployment, as well as adherence to responsible release practices such as gated model access and usage guidelines. We believe that advancing the scientific understanding of unified multimodal architectures carries substantial positive societal value, while the associated risks can be effectively managed through community-driven safety standards.

\section{Additional Main Results}
\label{appendix:additional}
This section provides the full per-category breakdown of the benchmarks summarised in Section~\ref{sec:exp}, complementing the condensed tables in the main paper.

\paragraph{GenEval.} Table~\ref{tab:geneval_detailed} reports the per-category breakdown on GenEval~\citep{ghosh2023geneval}, covering single-object, two-object, counting, color, position, and color-attribute prompts. The breakdown helps locate the compositional dimensions where each model is strongest.
\begin{table*}[!h]
    \centering
    \caption{\textbf{Detailed image generation results on the GenEval benchmark~\citep{ghosh2023geneval}.} Rows in gray indicate models with $\geq 14$B parameters and are excluded from the ranking. $^\dagger$ refers to methods using LLM rewriters.}
    \label{tab:geneval_detailed}
    \begin{adjustbox}{max width=0.95\linewidth}
    \begin{tabular}{@{}lc cccccc c@{}}
    \toprule
    \textbf{Models} & \textbf{Size} & \textbf{Single Object} & \textbf{Two Objects} & \textbf{Count} & \textbf{Colors} & \textbf{Position} & \textbf{Color Attribute} & \textbf{Overall} \\ \midrule
    
    \multicolumn{9}{l}{\textit{Generation-only Models}} \\
    \midrule
    SD3-Med \citep{esser2024scaling}             & 2B   & \textbf{0.99} & \textbf{0.94} & \underline{0.72} & \underline{0.89} & \underline{0.33} & \underline{0.60} & \underline{0.74} \\
    FLUX.1 [Dev] \citep{flux}                    & 12B  & \underline{0.98} & \underline{0.93} & \textbf{0.75}    & \textbf{0.93}    & \textbf{0.68}    & \textbf{0.65}    & \textbf{0.82}    \\ 
    DALL-E 3 \citep{betker2023dalle3}                       & --   & 0.96          & 0.87          & 0.47             & 0.83             & 0.43             & 0.45             & 0.67             \\ \midrule
    
    \multicolumn{9}{l}{\textit{Unified Multimodal Models}} \\
    \midrule
    \g{TokenFlow-XL \citep{qu2025tokenflow}}     & \g{14B} & \g{0.95} & \g{0.60} & \g{0.41} & \g{0.81} & \g{0.16} & \g{0.24} & \g{0.55} \\
    \g{SEED-X \citep{seed-x}}                    & \g{17B} & \g{0.97} & \g{0.58} & \g{0.26} & \g{0.80} & \g{0.19} & \g{0.14} & \g{0.49} \\
    \g{Ming-UniVision \citep{huang2025ming}}     & \g{16B} & \g{1.00} & \g{0.93} & \g{0.59} & \g{0.93} & \g{0.92} & \g{0.70} & \g{0.85} \\
    \g{BAGEL$^{\dagger}$ \citep{bagel}}          & \g{14B} & \g{0.98} & \g{0.95} & \g{0.84} & \g{0.95} & \g{0.78} & \g{0.77} & \g{0.88} \\
    MetaQuery-XL \citep{pan2025transfer}         & 7B   & --            & --            & --               & --               & --               & --               & 0.80             \\
    Blip3-o$^{\dagger}$ \citep{blip3-o}          & 8B   & --            & --            & --               & --               & --               & --               & 0.84             \\ 
    MUSE-VL \citep{muse-vl}                      & 7B   & 0.98          & 0.64          & 0.54             & 0.72             & 0.25             & 0.31             & 0.57             \\
    Janus-Pro \citep{chen2025janus}              & 7B   & \underline{0.99} & 0.89       & 0.59             & 0.90             & 0.79             & 0.66             & 0.80             \\
    Show-o2$^{\dagger}$ \citep{xie2025show}      & 7B   & \textbf{1.00} & 0.87          & 0.58             & \textbf{0.92}    & 0.52             & 0.62             & 0.76             \\
    HYDRA$^{\dagger}$ \citep{hydra}              & 7B   & \textbf{1.00} & \textbf{0.97} & \underline{0.68} & \underline{0.91} & \underline{0.81} & \textbf{0.80}    & \underline{0.86} \\ 
    \rowcolor{gray!10}
    \modelname                                  & 7B   & \underline{0.99} & \underline{0.95} & \textbf{0.83} & 0.90          & \textbf{0.84}    & \underline{0.75} & \textbf{0.88}    \\ 
    \bottomrule
    \end{tabular}
    \end{adjustbox}
\end{table*}

\paragraph{WISE.} Table~\ref{tab:wise_detailed} reports the per-category breakdown on WISE~\citep{niu2025wise}, which probes world knowledge across culture, time, space, biology, physics, and chemistry, and is therefore complementary to the geometric and compositional probes of GenEval.
\begin{table*}[!h]
    \centering
    \caption{\textbf{Detailed image generation results on the WISE benchmark~\citep{niu2025wise}.} Rows in gray indicate models with $\geq 14$B parameters and are excluded from the ranking.}
    \label{tab:wise_detailed}
    \begin{adjustbox}{max width=0.95\linewidth}
    \begin{tabular}{@{}lc ccccccc@{}}
    \toprule
    \textbf{Models} & \textbf{Size} & \textbf{Culture} & \textbf{Time} & \textbf{Space} & \textbf{Biology} & \textbf{Physics} & \textbf{Chemistry} & \textbf{Overall} \\ \midrule
    
    \multicolumn{9}{l}{\textit{Generation-only Models}} \\
    \midrule
    FLUX.1 [Dev] \citep{flux}                    & 12B  & \textbf{0.48}    & \textbf{0.58}    & \textbf{0.62}    & \underline{0.42} & \underline{0.51} & \textbf{0.35}    & \textbf{0.50}    \\
    SD3.5-Large \citep{esser2024scaling}         & 8B   & \underline{0.44} & \underline{0.50} & \underline{0.58} & \textbf{0.44}    & \textbf{0.52}    & \underline{0.31} & \underline{0.46} \\ \midrule
    
    \multicolumn{9}{l}{\textit{Unified Multimodal Models}} \\
    \midrule
    \g{BAGEL \citep{bagel}}                      & \g{14B} & \g{0.44} & \g{0.55} & \g{0.68} & \g{0.44} & \g{0.60} & \g{0.39} & \g{0.52} \\
    VILA-U-7B \citep{wu2024vila}                 & 7B   & 0.26             & 0.33             & 0.37             & 0.35             & 0.39             & 0.23             & 0.31             \\
    Janus-Pro-7B \citep{chen2025janus}           & 7B   & 0.30             & 0.37             & 0.49             & 0.36             & 0.42             & 0.26             & 0.35             \\
    Emu3-Gen-8B \citep{emu3}             & 8B   & 0.34             & 0.45             & 0.48             & 0.41             & 0.45             & 0.27             & 0.39             \\
    Show-o2 \citep{xie2025show}                  & 7B   & 0.40             & 0.45             & 0.58             & 0.39             & 0.53             & 0.34             & 0.44             \\
    HYDRA \citep{hydra}                          & 7B   & \underline{0.52} & \underline{0.53} & \underline{0.68} & \underline{0.47} & \underline{0.58} & \underline{0.38} & \underline{0.53} \\
    \rowcolor{gray!10}
    \modelname                                  & 7B   & \textbf{0.53}    & \textbf{0.57}    & \textbf{0.72}    & \textbf{0.53}    & \textbf{0.64}    & \textbf{0.40}    & \textbf{0.56}    \\ \bottomrule
    \end{tabular}
    \end{adjustbox}
\end{table*}

\paragraph{ImgEdit-Bench.} Table~\ref{tab:img_edit_detailed} provides the full per-dimension breakdown on ImgEdit-Bench~\citep{ye2025imgedit}, spanning nine instruction-guided editing operations from object addition and removal to background replacement, style transfer, and compositional edits.
\begin{table*}[!t]
    \centering
    \caption{\textbf{Detailed image editing results on the ImgEdit-Bench~\citep{ye2025imgedit}.} Editing dimensions: \textbf{Add}, \textbf{Adj.}\,(Alter), \textbf{Ext.}\,(Extract), \textbf{Rep.}\,(Replace), \textbf{Rm.}\,(Remove), \textbf{Bg.}\,(Background), \textbf{Sty.}\,(Style), \textbf{Hyb.}\,(Compose), \textbf{Act.}\,(Action). Rows in gray indicate models with $\geq 14$B parameters and are excluded from the ranking.}
    \label{tab:img_edit_detailed}
    \begin{adjustbox}{max width=0.95\linewidth}
    \begin{tabular}{@{}lcccccccccccc@{}}
    \toprule
    \textbf{Models} & \textbf{Size} & \textbf{Add} & \textbf{Adj.} & \textbf{Ext.} & \textbf{Rep.} & \textbf{Rm.} & \textbf{Bg.} & \textbf{Sty.} & \textbf{Hyb.} & \textbf{Act.} & \textbf{Overall} \\ \midrule
    \multicolumn{12}{l}{\textit{Generation-only Models}} \\
    \midrule
    FLUX.1 Kontext [Pro] \citep{flux}      & 12B & \underline{4.25} & \underline{4.15} & \underline{2.35} & \underline{4.56} & \underline{3.57} & \underline{4.26} & \underline{4.57} & \underline{3.68} & \underline{4.63} & \underline{4.00} \\
    Qwen-Image \citep{wu2025qwen-image}    & 20B & \textbf{4.38}    & \textbf{4.16}    & \textbf{3.43}    & \textbf{4.66}    & \textbf{4.14}    & \textbf{4.38}    & \textbf{4.81}    & \textbf{3.82}    & \textbf{4.69}    & \textbf{4.27}    \\ \midrule
    
    \multicolumn{12}{l}{\textit{Unified Multimodal Models}} \\
    \midrule
    \g{BAGEL \citep{bagel}}                 & \g{14B} & \g{3.56} & \g{3.31} & \g{1.70} & \g{3.30} & \g{2.62} & \g{3.24} & \g{4.49} & \g{2.38} & \g{4.17} & \g{3.20} \\
    OmniGen \citep{xiao2025omnigen}         & 3.8B  & 3.47             & 3.04             & 1.71             & 2.94             & 2.43             & 3.21             & 4.19             & 2.24             & 3.38             & 2.96             \\
    UniWorld-V1 \citep{lin2025uniworld}     & 12B   & \underline{3.82} & \underline{3.64} & \underline{2.27} & 3.47             & \underline{3.24} & 2.99             & 4.21             & \underline{2.96} & 2.74             & 3.26             \\
    OmniGen2 \citep{wu2025omnigen2}         & 4B    & 3.57             & 3.06             & 1.77             & \underline{3.74} & 3.20             & \underline{3.57} & \textbf{4.81}    & 2.52             & \textbf{4.68}    & \underline{3.44} \\
    \rowcolor{gray!10}
    \modelname                              & 7B    & \textbf{4.49}    & \textbf{4.27}    & \textbf{4.04}    & \textbf{4.41}    & \textbf{4.38}    & \textbf{4.30}    & \underline{4.77} & \textbf{3.43}    & \underline{4.32} & \textbf{4.34}    \\ \bottomrule
    \end{tabular}
    \end{adjustbox}
\end{table*}

\paragraph{VBench.} Table~\ref{tab:vbench} expands the QS/SS/Total summary in the main paper to all fourteen VBench~\citep{huang2024vbench} dimensions, separately probing visual quality, motion smoothness, dynamic degree, semantic correctness, and compositional reasoning.
\begin{table*}[!h]
    \centering
    \caption{\textbf{Detailed video generation results on VBench~\citep{huang2024vbench}.} Column abbreviations: \textbf{QS}: Quality Score, \textbf{SS}: Semantic Score, \textbf{SC}: Subject Consistency, \textbf{BC}: Background Consistency, \textbf{MS}: Motion Smoothness, \textbf{DD}: Dynamic Degree, \textbf{AQ}: Aesthetic Quality, \textbf{IQ}: Imaging Quality, \textbf{OC}: Object Class, \textbf{MO}: Multiple Objects, \textbf{HA}: Human Action, \textbf{C}: Color, \textbf{SR}: Spatial Relationship, \textbf{S}: Scene.}
    \vspace{-2mm}
    \label{tab:vbench}
    \begin{adjustbox}{max width=\linewidth}
    \scriptsize
    \begin{tabular}{@{}lc cccccccccccccc c@{}}
    \toprule
    \textbf{Models} & \textbf{Size} & \textbf{QS} & \textbf{SS} & \textbf{SC} & \textbf{BC} & \textbf{MS} & \textbf{DD} & \textbf{AQ} & \textbf{IQ} & \textbf{OC} & \textbf{MO} & \textbf{HA} & \textbf{C} & \textbf{SR} & \textbf{S} & \textbf{Total} \\ 
    \midrule
    \multicolumn{17}{l}{\textit{Generation-only Models}} \\
    \midrule
    CogVideoX \citep{yang2024cogvideox}        & 5B  & \underline{82.75} & \textbf{77.04}    & \underline{96.23} & \underline{96.52} & \underline{96.92} & \underline{70.97} & \textbf{61.98}    & \underline{62.90} & \textbf{85.23}    & \underline{62.11} & \textbf{99.40}    & \underline{82.81} & \underline{66.35} & \underline{53.20} & \underline{81.61} \\
    Hunyuan Video \citep{kong2024hunyuanvideo} & 13B & \textbf{85.07}    & \underline{76.88} & \textbf{97.22}    & \textbf{97.60}    & \textbf{99.05}    & \textbf{71.94}    & \underline{60.28} & \textbf{67.24}    & \underline{83.48} & \textbf{66.71}    & \underline{94.40} & \textbf{89.79}    & \textbf{72.13}    & \textbf{54.46}    & \textbf{83.43}    \\
    \midrule
    \multicolumn{17}{l}{\textit{Unified Multimodal Models}} \\
    \midrule
    VILA-U \citep{wu2024vila}        & 7B   & 76.26             & 65.04             & --                & --                & --                & --                & --                & --                & --                & --                & --                & --                & --                & --                & 74.01             \\
    HaploOmni \citep{xiao2025haploomni} & 7B & --                & --                & \underline{96.40} & \underline{97.60} & 96.80             & \underline{65.30} & --                & --                & --                & --                & --                & --                & --                & --                & 78.10             \\
    Emu3 \citep{emu3}                & 8B   & --                & --                & 95.32             & \textbf{97.69}    & \textbf{98.93}    & \textbf{79.27}    & 59.64             & --                & 86.17             & 44.64             & 77.71             & --                & \underline{68.73} & 37.11             & 80.96             \\
    Show-o2 \citep{xie2025show}      & 1.5B & \underline{82.10} & \underline{78.31} & \textbf{97.28}    & 96.78             & \underline{98.25} & 40.83             & \textbf{65.15}    & \textbf{67.06}    & \underline{94.81} & \underline{76.01} & \underline{95.20} & \underline{80.89} & 62.61             & \underline{57.67} & \underline{81.34} \\
    \rowcolor{gray!10}
    \modelname                       & 7B   & \textbf{83.97}    & \textbf{81.57}    & 95.70             & 95.88             & 92.69             & 35.42             & \underline{63.73} & \underline{65.23} & \textbf{96.52}    & \textbf{76.68}    & \textbf{99.00}    & \textbf{87.37}    & \textbf{73.88}    & \textbf{69.74}    & \textbf{83.49}    \\ 
    \bottomrule
    \end{tabular}
    \end{adjustbox}
    \vspace{-1em}
\end{table*}

\clearpage
\section{Qualitative Comparisons}
\label{appendix:qualitative}

This section presents qualitative results across the five tasks supported by \modelname. We compare against representative baselines drawn from both unified multimodal models and task-specialised systems, and organise the comparisons by task and resolution.

\subsection{\texorpdfstring{Image Reconstruction at $512{\times}512$}{Image Reconstruction at 512x512}}
\label{appendix:rec:512}

We first inspect reconstruction fidelity at the standard $512{\times}512$ resolution. The comparison spans three families of baselines: dedicated image VAEs (FLUX), unified tokenizers built into UMMs (MingTok, AToken), and the recently proposed RAE. The visual difference makes texture, fine-edge, and small-text fidelity directly comparable.

\begin{figure}[h]
    \centering
    \includegraphics[width=0.98\linewidth]{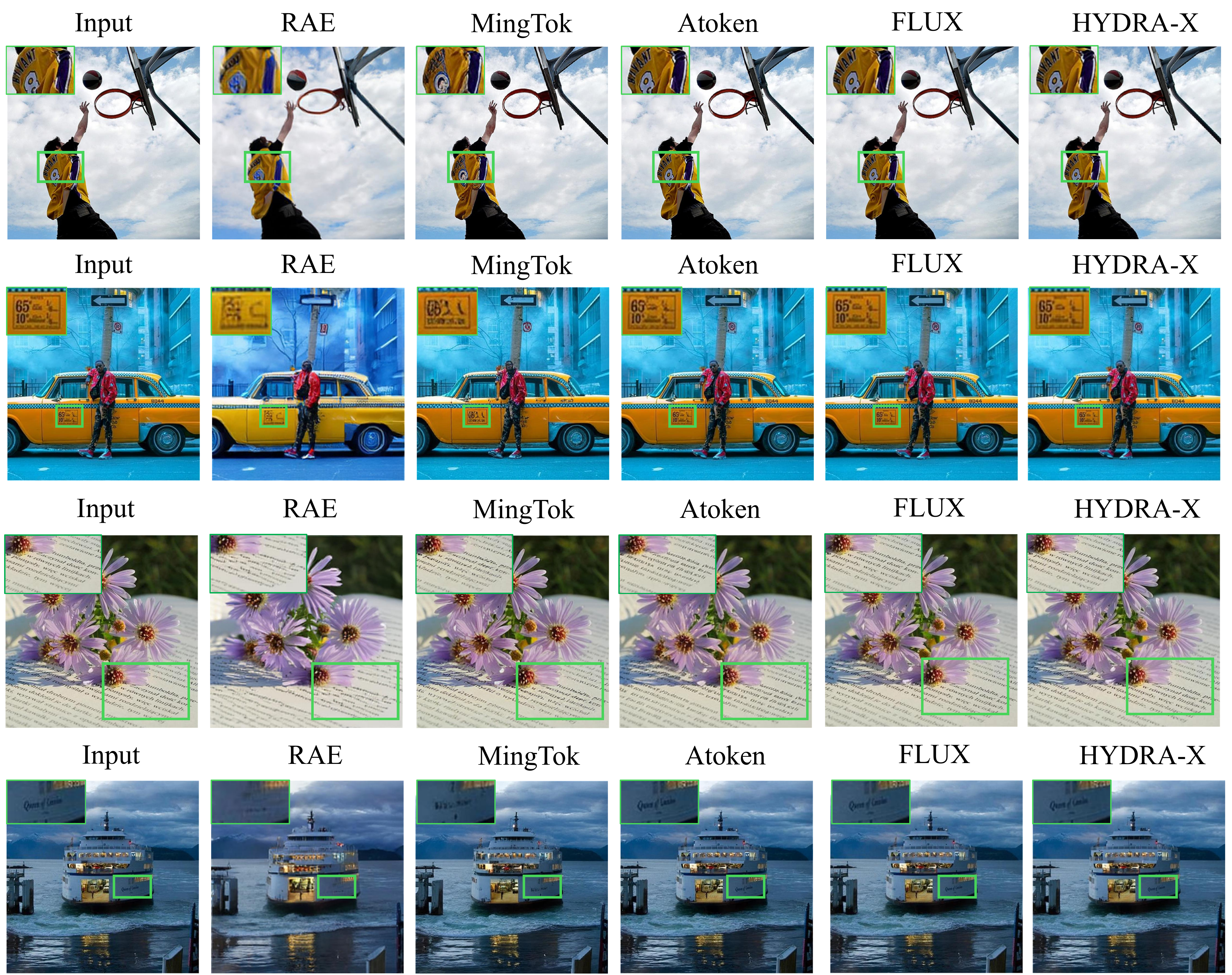}
    \caption{\textbf{Qualitative reconstruction comparison at $512{\times}512$.} We compare \modelname against RAE~\citep{rae}, MingTok~\citep{huang2025ming}, AToken~\citep{atoken}, and FLUX~\citep{flux}.}
    \label{fig:recon_compare_512}
\end{figure}

\clearpage
\subsection{\texorpdfstring{Image Reconstruction at $1280{\times}768$}{Image Reconstruction at 1280x768}}
\label{appendix:rec:1280}

To stress-test generalisation beyond the training resolution, we additionally compare reconstructions at a high resolution of $1280{\times}768$ and include the dedicated video VAE Wan~2.2 alongside the image-only baselines. This setting exposes how each tokenizer handles dense fine details such as text, foliage, and small structural elements when the spatial token budget is stretched.

\begin{figure}[h]
    \centering
    \includegraphics[width=0.98\linewidth]{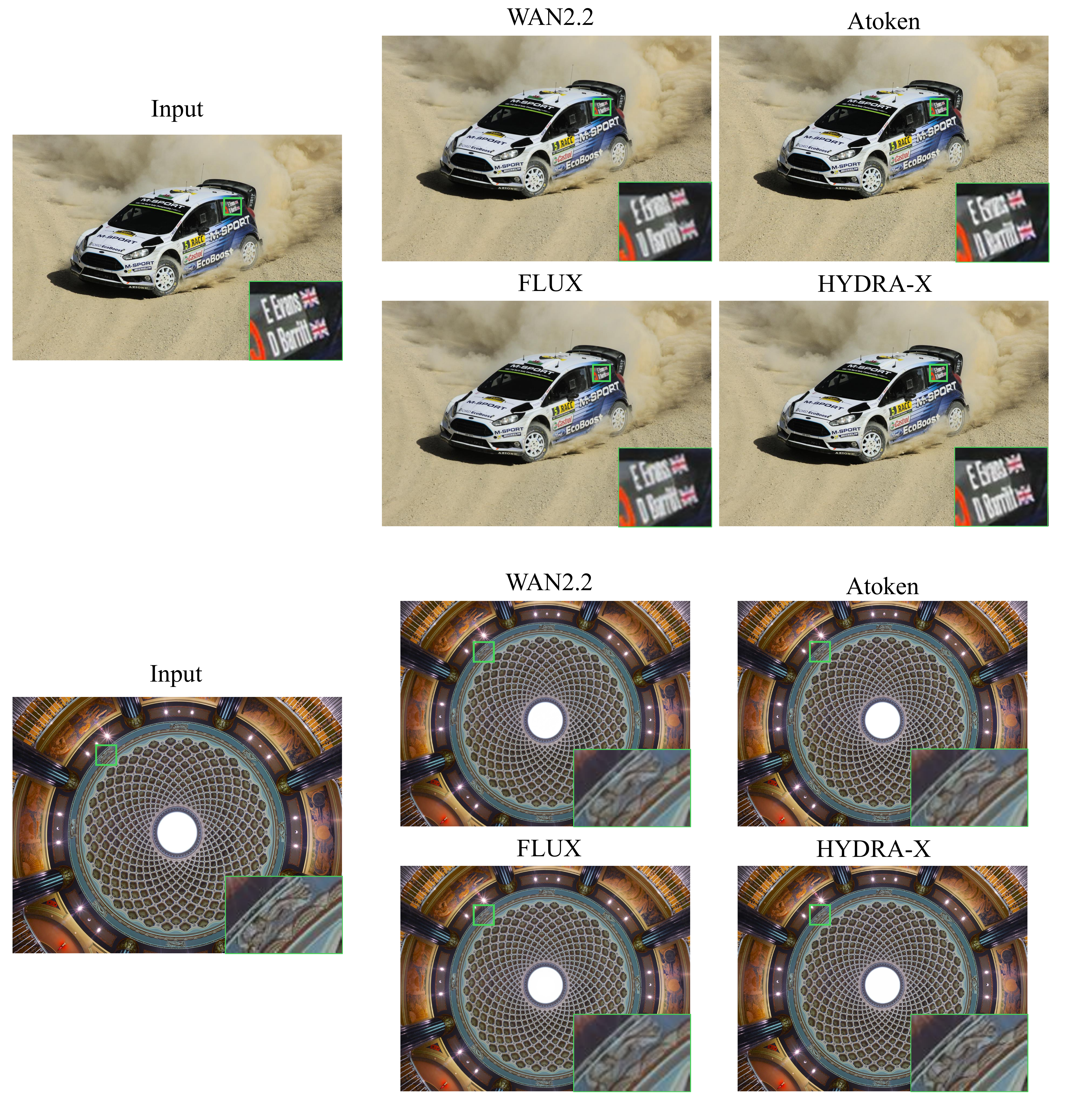}
    \caption{\textbf{Qualitative reconstruction comparison at $1280{\times}768$.} We compare \modelname against Wan~2.2~\citep{wan2.2}, AToken~\citep{atoken}, and FLUX~\citep{flux}.}
    \label{fig:recon_compare_1280}
\end{figure}

\clearpage
\subsection{\texorpdfstring{Video Reconstruction at $512{\times}512$}{Video Reconstruction at 512x512}}
\label{appendix:rec:video}

Beyond static images, we visualise temporally consecutive frames reconstructed by \modelname against the dedicated video VAE Wan~2.2 and the joint image--video tokenizer AToken. This helps assess whether \tokname's holistic ViT preserves motion-sensitive cues such as object boundaries and inter-frame consistency.

\begin{figure}[!htbp]
    \centering
    \includegraphics[width=0.55\linewidth]{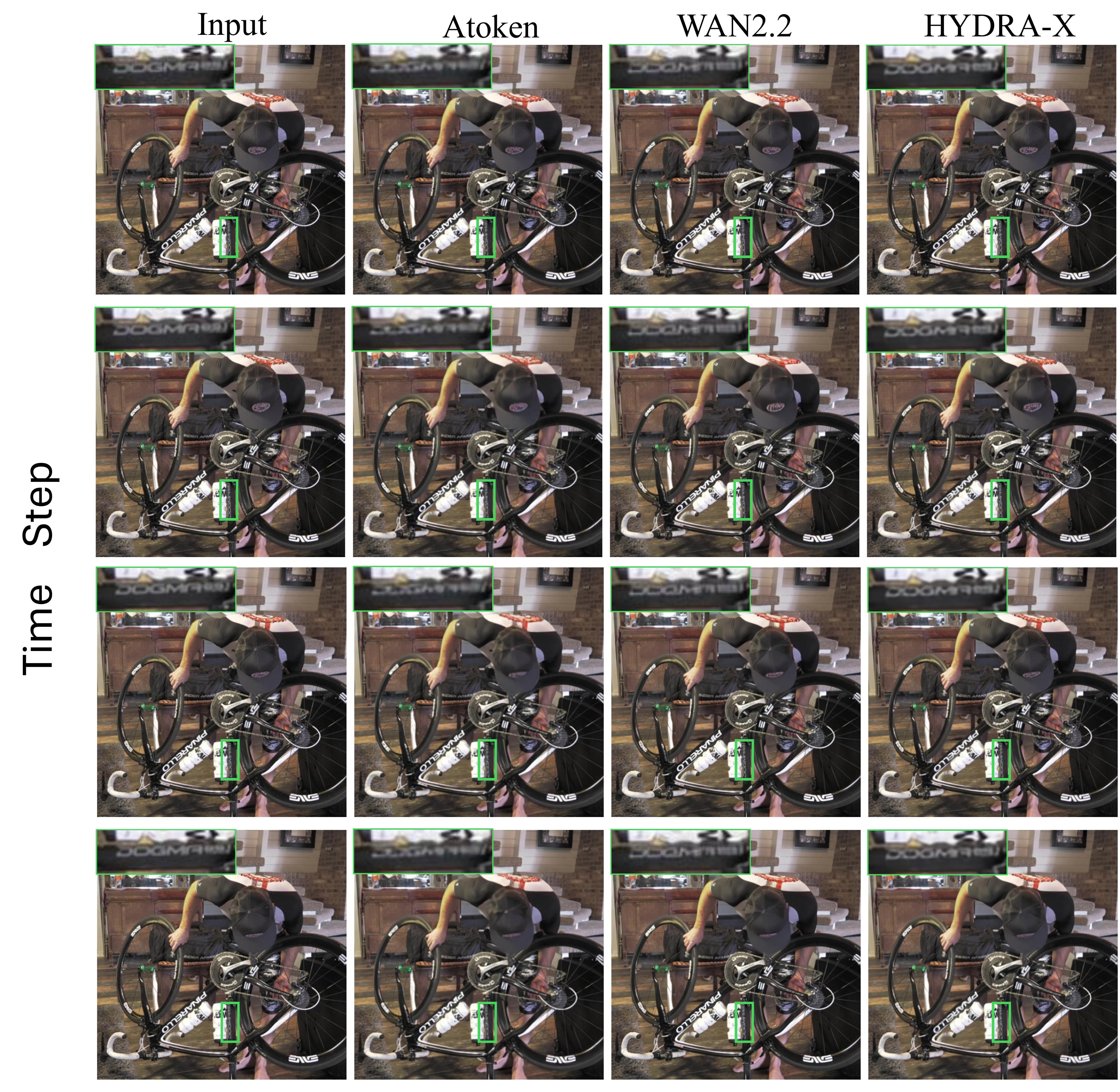}\\[2pt]
    \includegraphics[width=0.55\linewidth]{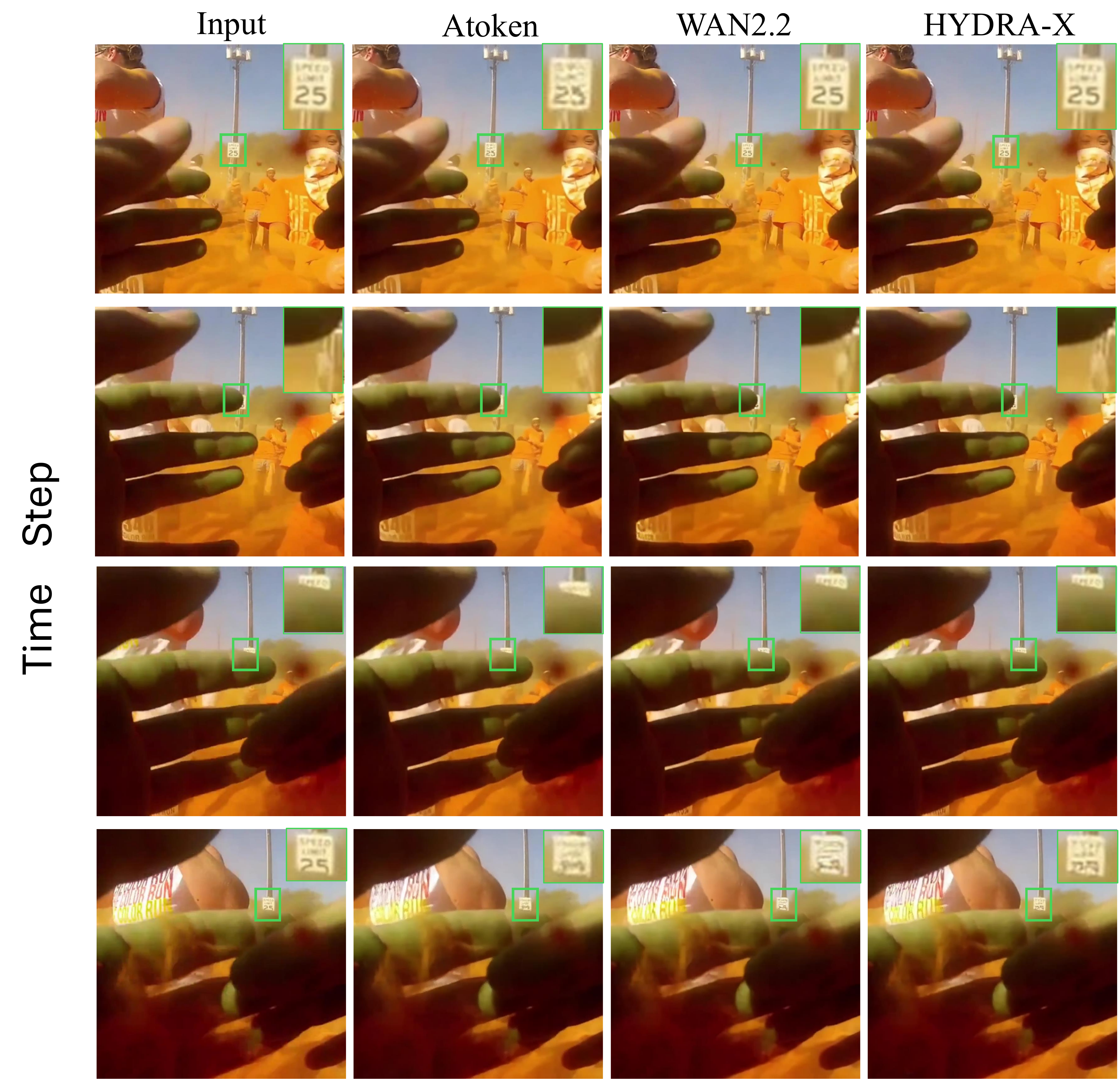}
    \caption{\textbf{Qualitative video reconstruction comparison.} We compare \modelname against Wan~2.2~\citep{wan2.2} and AToken~\citep{atoken}.}
    \label{fig:recon_compare_video}
\end{figure}

\clearpage
\subsection{Image Generation}
\label{appendix_t2i}

We provide qualitative text-to-image samples produced by \modelname spanning a diverse range of prompts---from realistic photography and stylised illustration to compositional and knowledge-driven scenes---to characterise the model's coverage and aesthetic quality.

\begin{figure}[h]
    \centering
    \includegraphics[width=0.90\linewidth]{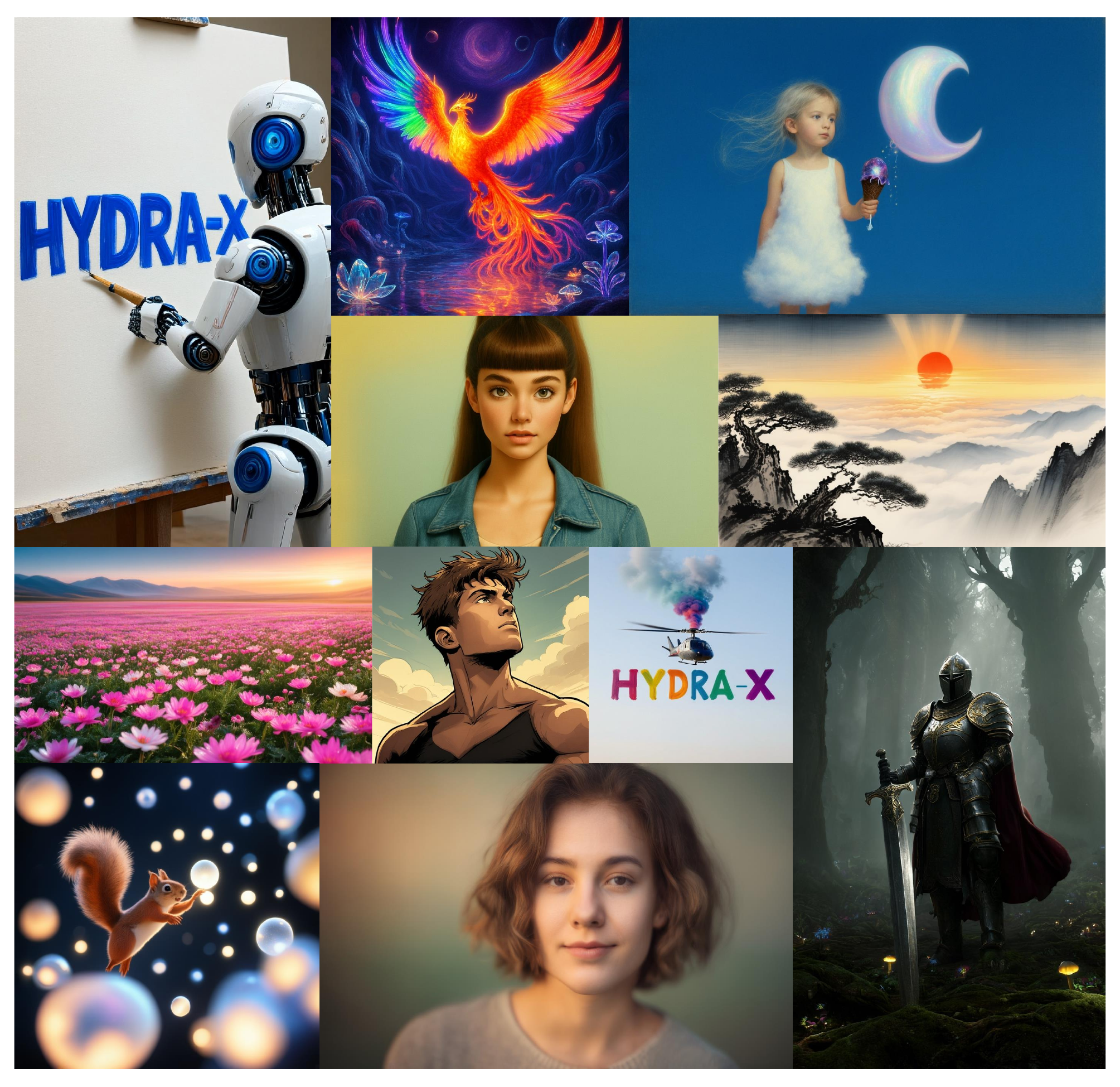}
    \caption{\textbf{Qualitative image generation results from \modelname.}}
    \label{fig:image_gen}
\end{figure}

\clearpage
\subsection{Video Generation}
\label{appendix:video_gen}

We similarly present qualitative text-to-video samples covering varied subjects, scenes, and motion patterns, illustrating how the holistic latent supports temporally coherent synthesis under the same UMM backbone.

\begin{figure}[h]
    \centering
    \includegraphics[width=0.90\linewidth]{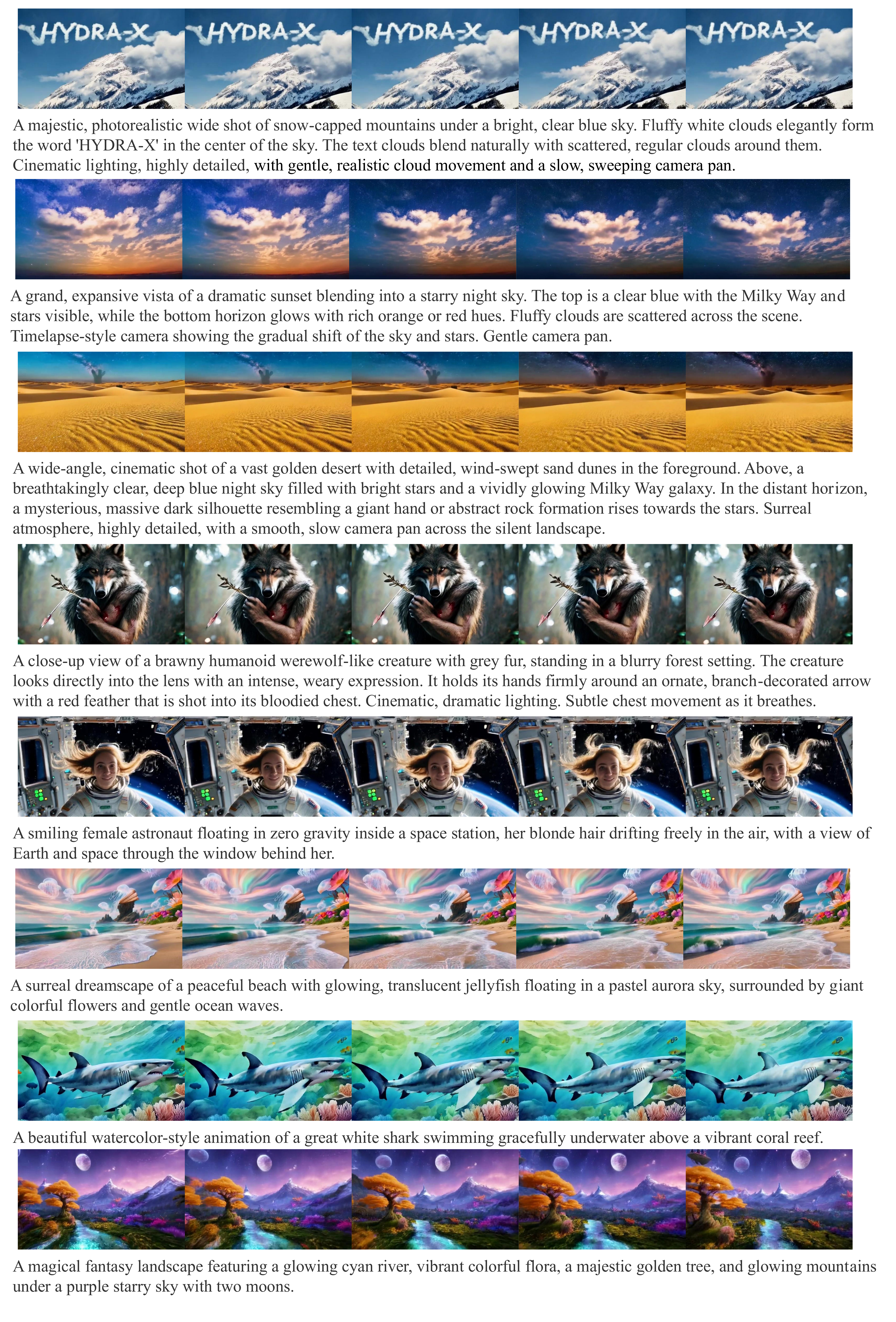}
    \caption{\textbf{Qualitative video generation results from \modelname.}}
    \label{fig:video_gen}
\end{figure}

\clearpage
\subsection{Image Editing}
\label{appendix:editing}

Finally, we compare \modelname against representative editing systems on a set of instruction-guided edits. The baselines include both unified multimodal models (BAGEL, OmniGen2) and editing-specialised generators (Qwen-Image-Edit, Step1X-Edit), allowing readers to gauge identity preservation, instruction adherence, and visual quality side-by-side.

\begin{figure}[h]
    \centering
    \includegraphics[width=0.90\linewidth]{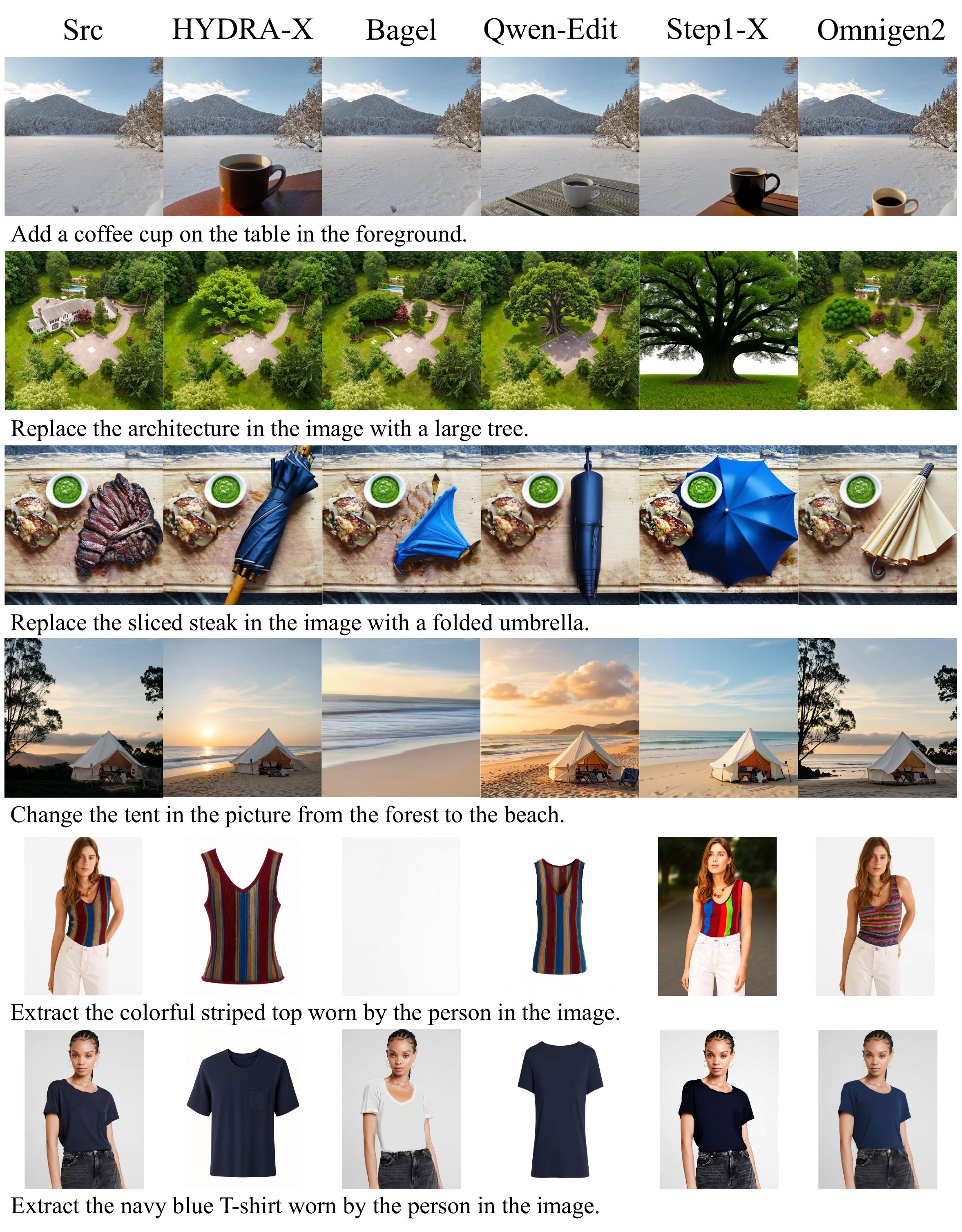}
    \caption{\textbf{Qualitative editing comparison.} We compare \modelname against BAGEL~\citep{bagel}, Qwen-Image-Edit~\citep{wu2025qwen-image}, Step1X-Edit~\citep{gedit}, and OmniGen2~\citep{wu2025omnigen2}.}
    \label{fig:edit_compare}
\end{figure}

\end{document}